\documentclass[journal]{IEEEtran}

\ifCLASSINFOpdf
\else
   \usepackage[dvips]{graphicx}
\fi
\usepackage{url}
\usepackage{graphicx}
\usepackage{cite}
\usepackage{amsmath,amssymb,amsfonts}
\usepackage{bm}
\usepackage{textcomp}
\usepackage{subfigure}
\usepackage{caption}
\usepackage{xcolor}
\usepackage{booktabs}
\usepackage{arydshln}
\usepackage{bbding}
\usepackage{multirow}
\usepackage{diagbox}
\usepackage{ntheorem}
\usepackage{makecell}
\usepackage{colortbl}
\usepackage{pifont}%
\usepackage{tablefootnote}
\usepackage[ruled,vlined]{algorithm2e}

\usepackage[pagebackref=true,breaklinks=true,colorlinks,bookmarks=false]{hyperref}

\usepackage[font={bf},textfont=md]{caption}

\usepackage[switch]{lineno} 
\usepackage{siunitx}

\newcommand{\figref}[1]{Fig.~\ref{#1}}

\newcommand{\secref}[1]{Sec.~\ref{#1}}
\newcommand{\tabref}[1]{Table.~\ref{#1}}

\newcommand{\algref}[1]{Alg.~\ref{#1}}

\def\X{\mathbf{X}}
\def\Y{\mathbf{Y}}
\def\G{\bm{\Gamma}}
\def\g{\bm{\gamma}}

\usepackage{xspace}
\makeatletter
\DeclareRobustCommand\onedot{\futurelet\@let@token\@onedot}
\def\@onedot{\ifx\@let@token.\else.\null\fi\xspace}

\def\eg{\emph{e.g}\onedot} 
\def\ie{\emph{i.e}\onedot} 
\def\cf{\emph{c.f}\onedot} 
\def\etc{\emph{etc}\onedot}

\makeatother

\definecolor{seagreen}{RGB}{84,255,159}
\definecolor{SpringGreen}{RGB}{0,139,69}

\newcolumntype{C}[1]{>{\PreserveBackslash\centering}p{#1}}
\newcolumntype{R}[1]{>{\PreserveBackslash\raggedleft}p{#1}}
\newcolumntype{L}[1]{>{\PreserveBackslash\raggedright}p{#1}}

\begin{document}

\title{Learning a Task-specific Descriptor for Robust Matching of 3D Point Clouds}

\author{Zhiyuan Zhang, Yuchao Dai \IEEEmembership{Member, IEEE}, Bin Fan, Jiadai Sun, and Mingyi He \IEEEmembership{Senior Member, IEEE}
\thanks{Zhiyuan Zhang, Yuchao Dai, Bin Fan, Jiadai Sun, and Mingyi He are with School of Electronics and Information, Northwestern Polytechnical University, China. Yuchao Dai (daiyuchao@gmail.com) is the corresponding author.}
}

\markboth{Journal of \LaTeX\ Class Files, Vol. 14, No. 8, August 2015}
{Shell \MakeLowercase{\textit{et al.}}: Bare Demo of IEEEtran.cls for IEEE Journals}

\maketitle

\makeatletter

\begin{abstract}
Existing learning-based point feature descriptors are usually \emph{task-agnostic}, which pursue describing the individual 3D point clouds as accurate as possible. 
However, the matching task aims at describing the corresponding points consistently across different 3D point clouds. Therefore these too accurate features may play a counterproductive role due to the inconsistent point feature representations of correspondences caused by the unpredictable noise, partiality, deformation, \etc, in the local geometry.
In this paper, we propose to learn a robust \emph{task-specific} feature descriptor to consistently describe the correct point correspondence under interference.
Born with an \underline{E}ncoder and a \underline{D}ynamic \underline{F}usion module, our method EDFNet develops from two aspects.
First, we augment the matchability of correspondences by utilizing their repetitive local structure. To this end, a special encoder is designed to exploit two input point clouds jointly for each point descriptor. It not only captures the local geometry of each point in the current point cloud by convolution, but also exploits the repetitive structure from paired point cloud by Transformer.
Second, we propose a dynamical fusion module to jointly use different scale features. 
There is an inevitable struggle between robustness and discriminativeness of the single scale feature. Specifically, the small scale feature is robust since little interference exists in this small receptive field. But it is not sufficiently discriminative as there are many repetitive local structures within a point cloud. Thus the resultant descriptors will lead to many incorrect matches. In contrast, the large scale feature is more discriminative by integrating more neighborhood information. But it is easier to be disturbed since there is much more interference in the large receptive field.
Compared with the conventional fusion strategy that handles multiple scale features equally, we analyze the consistency of them to judge the clean ones and perform larger aggregation weights on them during fusion.
Then, a robust and discriminative feature descriptor is achieved by focusing on multiple clean scale features.
Extensive evaluations validate that EDFNet learns a task-specific descriptor, which achieves state-of-the-art or comparable performance for robust matching of 3D point clouds.

\end{abstract}


\IEEEpeerreviewmaketitle

\section{Introduction} \label{sec::introduction}

\begin{figure}[t]
	\centerline{\includegraphics[width=\linewidth]{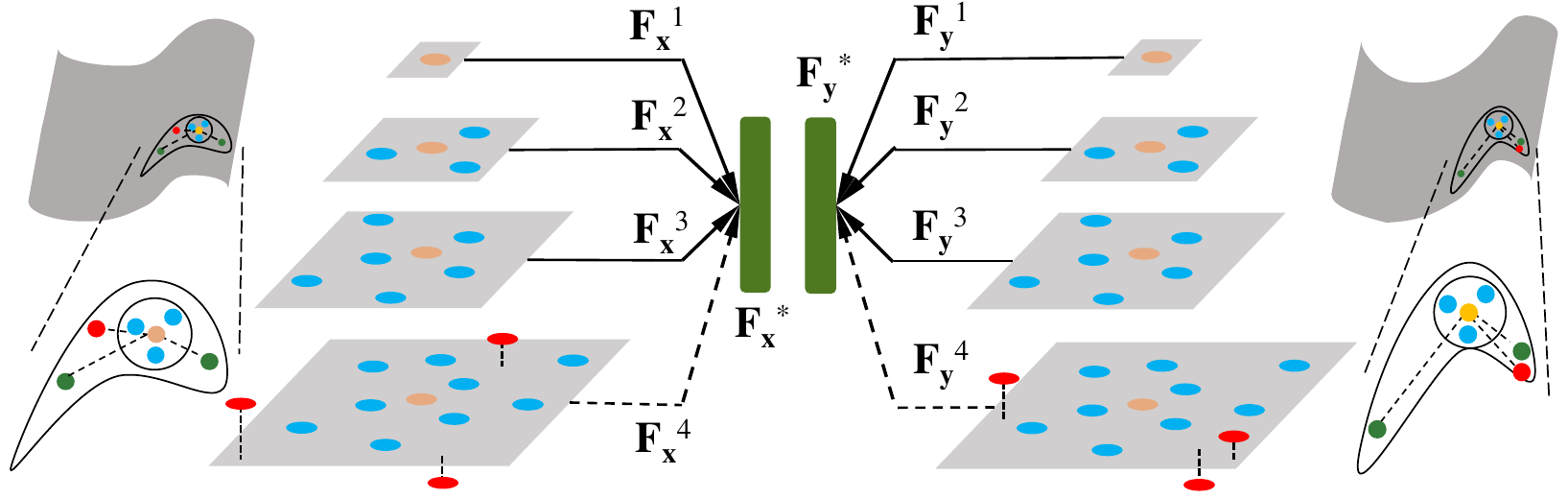}}
    \caption{Motivation of our proposed dynamic fusion strategy. There is an inevitable struggle between robustness and discriminativeness of single scale feature description. Due to repetitive minor structures, the small scale descriptors are not discriminative enough but are robust due to little interference. By contrast, the large receptive field improves the discriminativeness of large scale feature, which is easy to be affected by many disturbers. 
    Existing multi-scale fusion methods usually utilize each scale equally \cite{wang_dgcnn_tog_2019,charles_pointnet2_nips_2017,thomas_kpconv_iccv_19,D3Feat_bai_cvpr_2020}, which cannot improve the robustness as the disturbers are also encoded to the final descriptor.
    Thus, we design a dynamic fusion module, which recognizes and selects consistent and clean ones dynamically during the fusion. By focusing on multiple clean scale features, it achieves a good balance between discriminativeness and robustness.
    Here, we provide a toy example, where yellow denotes the current point (\ie $\mathbf{x}$, $\mathbf{y}$) and red is noise, gray parallelograms indicate different receptive fields. Obviously, the features of the first 3 scales $\mathbf{F}_\mathbf{x}^1$,$\mathbf{F}_\mathbf{x}^2$,$\mathbf{F}_\mathbf{x}^3$ and $\mathbf{F}_\mathbf{y}^1$,$\mathbf{F}_\mathbf{y}^2$,$\mathbf{F}_\mathbf{y}^3$ are consistent in describing similar local geometry (a plane). However, the last scale features $\mathbf{F}_\mathbf{x}^4$ and $\mathbf{F}_\mathbf{y}^4$ show deviation due to the noise and deformation. Our designed dynamic fusion module enforces the final feature $\mathbf{F}_\mathbf{x}^{*}$, $\mathbf{F}_\mathbf{y}^{*}$ approach to the first 3 scale features while marginalizing the last scale feature. }
	\label{fig:ill}
    \vspace{-0.7cm}
\end{figure}

\IEEEPARstart{W}{ith} the vigorous development of 3D measurement technology, point cloud has received widespread research attention \cite{wu_subjective_tcsvt_2021,qian_deep_tip_2021,liu_hybrid_tcsvt_2021}. Point matching plays fundamental roles in various applications, \eg 3D reconstruction \cite{deschaud_imls_icra_2018, agarwal2011building}, object pose estimation \cite{ming_segicp_iros_2017}, autonomous driving \cite{ huang_ctf_tcsvt_2018,ding_deepmapping_cvpr_2019}, \etc.
To achieve accurate matching, using the point feature descriptors to associate the source and target point clouds has been a widely adopted strategy. Many point feature descriptors have been proposed recently.

Before the era of deep learning, handcrafted point feature descriptors are always custom-tailored by investigating the neighborhood distribution of each 3D point (\eg \cite{andrew_usi_pami_1999,flintDH_thrift_cs_07,rusu_fpfh_icra_2009,SHOT_Tombari_2010_ECCV}). 
Recently, with the flourishing of deep learning, the exploration of feature descriptors has shifted to learning-based approaches, and superior performance has been achieved. Many methods use multi-layer perceptron (MLP) to handle the irregular point clouds to learn point feature descriptors \cite{charles_pointnet_cvpr_2017,charles_pointnet2_nips_2017,yew_3dfeatnet_eccv_2018}. 
Furthermore, convolution is applied \cite{choy_fcgf_iccv_2019,thomas_kpconv_iccv_19} to incorporate local information more effectively.
Besides, customized rotation-equivariant/-invariant descriptors \cite{esteves_equ_eccv_2018,rao_sfc_cvpr_2019,Fang_rotpredictor_3dv_2020,perfect_zan_cvpr_2019,PPFNet_Deng_2018_CVPR,PPFFolding_Deng_2018_ECCV} have been developed to handle the pose change.
However, in these existing methods, the obtained local features are always \emph{task-agnostic}, where feature extraction and matching are totally separated. They are committed to describing each 3D point as accurate as possible. Note that the matching task resorts to describing the corresponding points consistently. 
Then, these accurate feature descriptors may backfire since the local geometry is usually disturbed by noise, partiality, deformation, \etc. Then, the descriptors of corresponding points are thus inconsistent.

To resolve this problem, we propose a method termed EDFNet (\underline{E}ncoder-\underline{D}ynamic \underline{F}usion structure) to learn a \emph{task-specific} descriptor in this paper. Our descriptor is devoted to consistently describing the correspondence under irregular and unpredictable interference. EDFNet is developed from the following two aspects.

First, since the matching task solves for the point correspondences from the source and target point clouds, we advocate facilitating the matchability of correspondence based on their repetitive local structure. To this end, a \underline{c}onvolution and \underline{T}ransformer encoder (notated as CT-encoder) is designed. It not only learns the local geometry of each point in the current point cloud by convolution but also exploits the repetitive structure in the paired point cloud by using the Transformer \cite{vaswani_attention_nips_2017,superglue_paul_cvpr_20,LoFTR_zhou_cvpr_21}.
Different from \cite{wang_dcp_iccv_2019,wang_prnet_nips_2019,zhang_vrnet_tcsvt_2022} that use Transformer to sense the paired point cloud once, our CT-encoder strengthens the matchability of correspondence more sufficiently. Specifically, we apply multiple encoder layers to integrate the local and paired geometries from multiple scales.

Second, we observe that there exists an inevitable struggle between robustness and discriminativeness of single scale feature. 
Small scale features are not sufficiently discriminative because their small receptive fields cannot capture enough local geometry information. Then, many repetitive minor structures in point clouds will cause incorrect matches.
Nevertheless, this small scale feature is robust, which describes correspondence consistently due to little interference in such small receptive field. By contrast, the large receptive field improves the discriminativeness of large scale feature by encoding more neighborhood information. However, it is easy to be affected by embedding significantly increased disturbers as shown in \figref{fig:ill}. To achieve better matching, we advocate learning a robust and discriminative descriptor.
To this end, we propose to jointly use different scale features. 
However, we notice that the existing multiple scale feature fusion methods cannot handle the disturbers. These methods usually take addition \cite{thomas_kpconv_iccv_19,D3Feat_bai_cvpr_2020} or concatenation \cite{wang_dgcnn_tog_2019,charles_pointnet2_nips_2017,li_multifusion_tcsvt_2021} strategies, which essentially treat different scale features equally. Thus the contaminated scale features also contribute to the final fused descriptor.
Inspired by the fusion on 2D image investigation \cite{zhang_attention_tcsvt_2021,Fang_featurefusion_tcsvt_2021}, we propose a dynamic fusion module. 
Specifically, we observe an intuitive rule, \ie the clean scale features are usually consistent to describe the similar local geometries. And the contaminated ones are usually diverse due to the irregular disturbers. 
Thus, we propose to guide the obtained descriptor to approach to the consistent and clean scale features and keep away from the contaminated ones. 
To this end, we analyze the distribution of them and focus on the clean ones more during fusion. 
Through our dynamic fusion, the final obtained feature descriptor is robust and discriminative via utilizing multiple clean scale features.

With the above two strategies, our EDFNet achieves a robust matching task-specific point feature descriptor.
Extensive evaluations on several benchmark datasets have validated our method, which achieves a state-of-the-art \textit{(SOTA)} or comparable matching performance. Besides, the strong robustness and generalization ability of EDFNet have also been verified on rigid and non-rigid point cloud data.

Our contributions can be summarized as follows:
\begin{itemize}
\setlength{\itemsep}{0pt}
\setlength{\parsep}{0pt}
\setlength{\parskip}{0pt}
	\item We propose EDFNet to learn matching task-specific feature descriptors to consistently describe the correspondences under unpredictable interference. 
	\item We propose an encoder to extract multiple scale features by sensing both two input point clouds simultaneously, which consists of a convolution branch and a Transformer branch. Besides, we design a dynamic fusion module to guide the final fused feature to approach to the consistent and clean scale features and far away from the contaminated ones to reply to disturbers. This dynamic fusion module achieves a good balance between robustness and discriminativeness.
	\item Extensive experiments on several benchmark datasets show that our EDFNet achieves comparable if not \textit{SOTA} performance for robust 3D point cloud matching. 
\end{itemize}

\section{Related Work}\label{sec::relatedwork}
In this section, we briefly review the representative feature descriptors of 3D point clouds, including the handcrafted ones and the learning-based ones.

\noindent\textbf{Handcrafted Point Feature Descriptors.} 
To match 3D points in the source and target point clouds, many customized pose-invariant feature descriptors have been designed following two representative strategies, \ie local reference frame (LRF)-based approaches and LRF-free approaches. 
For the LRF-based approaches, the unique LRF is explored to transform the local neighborhood of 3D point to a canonical representation \cite{SHOT_Tombari_2010_ECCV,TOLDI_Yang_PR_2017,USC_Tombari_10,RoPS_Guo_2013_ijcv}. 
On the other hand, the LRF-free approaches are another hot spot, and many methods have emerged, \eg \cite{rusu_fpfh_icra_2009,PPF_Birdal_3dv_2015,APCV_Rusu_iros_2008}. 
These LRF-free methods usually focus on the intrinsically invariant geometry peculiarities.
Although significant progress has been achieved, the performance of these handcrafted descriptors based methods is limited. Furthermore, most existing handcrafted features cannot handle the noise, occlusions, deformation, \etc, \cite{Summary_Guo_IJCV_2016}.

\noindent\textbf{Learning-based Point Feature Descriptors.} 
Deep learning technology has inspired various 3D point cloud descriptor learning methods.
The early works apply deep learning after converting the raw irregular 3D point cloud data into regular data, such as 2D image \cite{3DAE_Elbaz_CVPR_2017,MCN_Huang_TOG_2018}, 3D voxel \cite{voxnet_maturana_iros_2015,shapenet_wu_cvpr_15,mvc_qi_cvpr_2016,3dmatch_zeng_cvpr_2017,choy_fcgf_iccv_2019}, \etc.
Recently, using deep learning to directly handle 3D unstructured point cloud data has become popular. To this end, PointNet \cite{charles_pointnet_cvpr_2017} introduces a symmetric function, max-pooling for permutation invariance. 
Then, to achieve a more descriptive descriptor, DGCNN \cite{wang_dgcnn_tog_2019} uses graph convolution to collect neighborhood information. 
PPFNet \cite{PPFNet_Deng_2018_CVPR} improves the feature descriptor by integrating the point pair feature.
3DFeat-Net \cite{yew_3dfeatnet_eccv_2018} uses the set abstraction module proposed in \cite{charles_pointnet2_nips_2017} to solve the feature descriptor, which learns the keypoint detector and descriptor simultaneously following \cite{LiFt_Yi_eccv_2016}. 
Besides, kernel convolution network is usually employed to better aggregate the local information to learn the point feature descriptor. PointConv \cite{wu_pointconv_cvpr_2019} designs an approximate convolution using MLP.
A set of kernel points is used to define the convolution area in KPConv \cite{thomas_kpconv_iccv_19}. 
D3Feat \cite{D3Feat_bai_cvpr_2020} achieves a density-invariant convolution by considering the whole supportive neighbors.

In addition, many learning-based rotation-equivariant and rotation-invariant descriptors have also received widespread attention.
For rotational equivariance, the point cloud is mapped into the spherical functions, and an equivariant spherical convolution operator is introduced in \cite{esteves_equ_eccv_2018,rao_sfc_cvpr_2019}. RotPredictor \cite{Fang_rotpredictor_3dv_2020} represents the rotations as translations and then applies a translation-equivariant convolution to achieve rotation-equivariance. To achieve rotation invariance, a rotation-invariant convolution kernel is designed in \cite{Poulenard_eri_3dv_2019} based on spherical harmonics. 
There are also some methods that apply networks to handcrafted rotation-invariant features (\eg relative
locations, distances, angles) for rotation invariance. 
RIConv \cite{zhang_roi_3dv_2019}, Triangle-Net \cite{xiao_triangle_wacv_2021}, PPF-FoldNet \cite{PPFFolding_Deng_2018_ECCV}, ClusterNet \cite{chen_clusternet_cvpr_2019}, and SRI-Net \cite{sun_srinet_acmmm_2019} are representative methods following this pipeline. Learning-based LRF-based methods \cite{zhang_lri_3dv_2020,xiao_endowing_icme_2020,zhao_quaternion_eccv_2020} also receive increasing attention.

Arguably, although the development of deep learning has encouraged many advanced feature descriptors, most of them are task-agnostic and easy to be disturbed.
In response, we propose EDFNet to learn a robust matching task-specific feature descriptor, which describes the correct correspondence consistently under interference.

\section{Method}\label{sec::method}

\begin{figure*}[!t]
	\centerline{\includegraphics[width=0.85\linewidth]{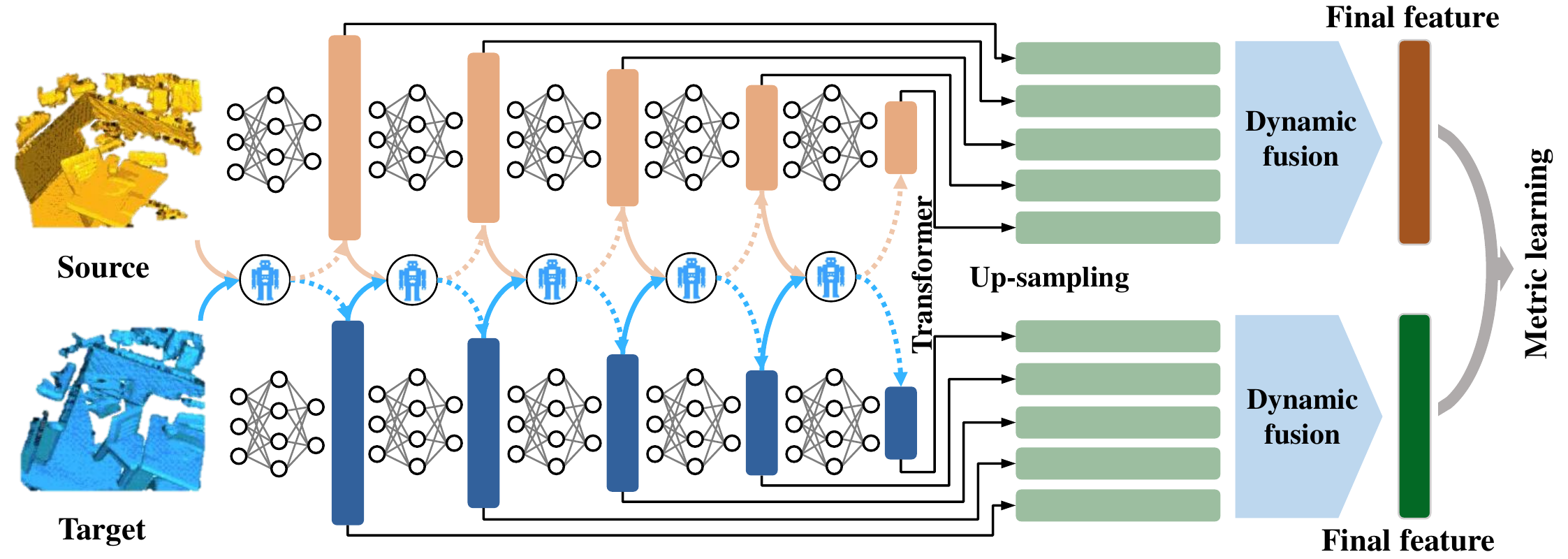}}
	\vspace{-0.3cm}
	\caption{Illustration of our proposed EDFNet architecture. Given two point clouds, the CT-encoder network is used to extract different scale features, where convolution and Transformer are employed to extract local information of the current point cloud and the repetitive structure information of paired point clouds, respectively. Then, we design a dynamic fusion module to fuse these different scale features to the final feature descriptor. The entire network is trained with a metric learning loss. Note that the robot indicates the Transformer module.}
	\label{Fig:network}
	\vspace{-16pt}
\end{figure*}

In this section, we introduce our proposed EDFNet to learn the point feature descriptors for robust matching of 3D point clouds. Note that existing learning-based feature descriptors are usually task-agnostic focusing on accurately describing 3D points. However, as mentioned above, these too accurate feature descriptors may play a counterproductive role for the matching task since the disturbers of noise, deformation, partiality, \etc, may change the local geometry and thus lead to inconsistent descriptors of correspondence. In response, we learn a matching task-specific point feature descriptor, which describes the correspondence consistently under the irregular and unpredictable interference.
To this end, a new deep learning pipeline named EDFNet is presented (\cf \figref{Fig:network}), which mainly consists of an encoder and a dynamic fusion module. In the encoder, we consider both the source and target point clouds synchronously to facilitate the matchability of correspondence by exploiting the repetitive structure during extracting different scale features. Moreover, different from the traditional encoder-decoder structure, EDFNet uses a dynamic fusion module instead of a decoder network to fuse these different scale features selectively to the final descriptor. To achieve a robust and discriminative descriptor, EDFNet explores the consistency of these different scale features themselves and guides the final descriptor to approach to consistent and clean ones and dodge the contaminated ones.

\subsection{CT-Encoder}
\label{sec::ED}
In this section, we use an encoder network to extract different scale features. 
Here, we exploit the repetitive structure in the input two point clouds to facilitate the matchability of correspondences. We learn the point features by capturing the local geometry information of the current point cloud and the relevant co-contextual information from the paired point cloud. Note that convolution \cite{D3Feat_bai_cvpr_2020} has been proved to be effective for local feature extraction and the Transformer \cite{vaswani_attention_nips_2017} has been proved to be suitable for global search. We propose a \underline{C}onvolution and \underline{T}ransformer encoder (CT-encoder), where each encoder layer consists of a convolution branch and a Transformer branch.
By applying our CT-encoder, different scale point features are learned, where the receptive field of each point feature is enlarged by summarizing the output of the previous layer. The size of the input point set of each encoder layer is reduced by down-sampling the input point set of the previous encoder layer. This operation is memory efficient in implementation in response to the huge memory consumption of the Transformer. Here, we take one encoder layer as an example for a clear explanation.

\noindent\textbf{Convolution branch.}
The convolution-based 3D point cloud processing \cite{thomas_kpconv_iccv_19,D3Feat_bai_cvpr_2020} has shown advanced performance in capturing the local geometry. It not only defines the convolution area according to the kernel points, but also achieves density invariance by considering all supporting neighbors. We build our convolution branch by adopting this convolution operation to capture local geometry.

Formally, the input of the $\ell$-th encoder layer includes the point set $\mathbf{X}^{\ell}=[\mathbf{x}_i^{\ell}]\in \mathbb{R}^{N^{\ell}\times 3}$, the corresponding point feature set $\mathbf{F}^{\ell}=[\mathbf{f}_{\mathbf{x}_i^\ell}] \in \mathbb{R}^{N^{\ell}\times D^{\ell}}$, and the point set of next layer $\mathbf{X}^{\ell+1}=[\mathbf{x}_i^{\ell+1}]\in \mathbb{R}^{N^{\ell+1}\times 3}$, where $\mathbf{X}^{\ell+1}\subset \mathbf{X}^{\ell}$, $N^{\ell}$ and $N^{\ell+1}$ indicate the sizes of the point sets, $D^{\ell}$ is the feature dimension of the $\ell$-th encoder layer. Note that the initial point feature is the 3D coordinate. Meanwhile, a density normalization term considering the whole supporting neighbor set is applied to ensure sparsity invariance. Then, the convolution feature of the next layer $\psi_{\mathbf{x}_i^{\ell+1}}$ at the point $\mathbf{x}_i^{\ell+1}\in \mathbf{X}^{\ell+1}$ is obtained from the convolution branch using the kernel $g$,
\begin{equation}
    \psi_{\mathbf{x}_i^{\ell+1}}=\frac{1}{|\mathcal{N}_{\mathbf{x}_i^{\ell+1}}|}\sum\nolimits_{\mathbf{x}_j^{\ell} \in \mathcal{N}_{\mathbf{x}_i^{\ell+1}}} g(\mathbf{x}_j^{\ell}-\mathbf{x}_i^{\ell+1})\mathbf{f}_{\mathbf{x}_j^{\ell}},
\end{equation}
where $\mathcal{N}_{\mathbf{x}_i^{\ell+1}}$ is the radius neighborhood of point ${\mathbf{x}_i^{\ell+1}}$ in the current point set, \ie $\mathcal{N}_{\mathbf{x}_i^{\ell+1}}\!=\!\{\mathbf{x}_j^\ell \!\in\! \mathbf{X}^{\ell}\ |\ \|\mathbf{x}_j^\ell - \mathbf{x}_i^{\ell+1}  \| \!\le\! \tau^{\ell}\}$, where $\tau^{\ell}$ is a pre-defined radius.
The kernel function $g(\cdot)$ is defined as $g(\mathbf{x}_j^{\ell}-\mathbf{x}_i^{\ell+1})=\sum_{s=1}^S h(\mathbf{x}_j^{\ell}-\mathbf{x}_i^{\ell+1}, \hat{\mathbf{x}}_s^{\ell})\mathbf{W}_s^{\ell}$,
where $h(\mathbf{x}_j^{\ell}-\mathbf{x}_i^{\ell+1}, \hat{\mathbf{x}}_s^{\ell})\!=\!\text{max}(0, 1-\|(\mathbf{x}_j^\ell-\mathbf{x}_i^{\ell+1}) - \hat{\mathbf{x}}_s^{\ell}\|/\sigma)$ is the correlation function between the kernel point $\hat{\mathbf{x}}_s^{\ell}$ and the supporting point $\mathbf{x}_j^\ell$, $\sigma$ is a hyperparameter. $\mathbf{W}_s^{\ell}$ is the weight matrix of the kernel point $\hat{\mathbf{x}}_s^{\ell}$, and $S$ is the number of kernel points. More details can be seen in \cite{D3Feat_bai_cvpr_2020,thomas_kpconv_iccv_19}. We denote the set of the obtained feature as $\Psi_{\mathbf{X}^{\ell+1}}=[\psi_{\mathbf{x}_i^{\ell+1}}]\in \mathbb{R}^{N^{\ell+1} \times D^{\ell+1}}$, and $D^{\ell+1}$ is the feature dimension of the next encoder layer.

\begin{figure}[!h]
	\centerline{\includegraphics[width=0.8\linewidth]{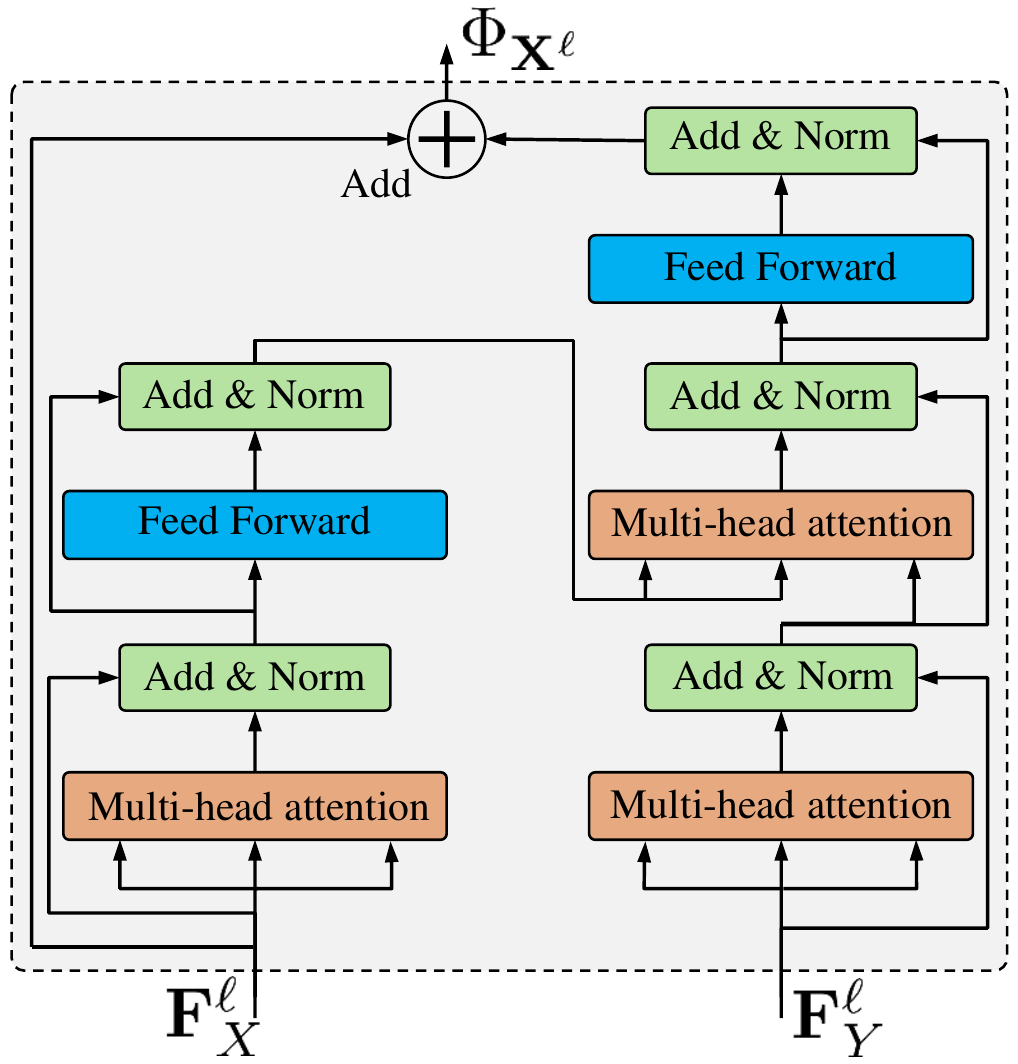}}
	\vspace{-0.1cm}
	\caption{The architecture of the Transformer branch. Two inputs correspond to the feature maps of source and target respectively. At first, the input feature is embedded by MLP to the query, key and value respectively. The skip connection structure is applied by adding the processed feature map to the original feature map, and the results are normalized further. The feed forward module is implemented by the MLP. By the interaction between the source and target, the matchability is augmented by the searched repetitive local structures.
	}
	\label{Fig:transformer}
	\vspace{-0.5cm}
\end{figure}

\noindent\textbf{Transformer branch.}
Besides learning the point feature by focusing on the source or target point cloud individually, we utilize Transformer \cite{vaswani_attention_nips_2017} to construct our Transformer branch in each encoder layer to explore the repetitive structure from the paired point clouds. The architecture of this branch is shown in \figref{Fig:transformer}, and refer to \cite{vaswani_attention_nips_2017} for more details.

As mentioned above, the features of the source and target point clouds at the current encoder layer are notated as $\mathbf{F}_{X}^{\ell}\!\in\! \mathbb{R}^{N_X^{\ell}\times D^{\ell}}$ and $\mathbf{F}_{Y}^{\ell}\!\in\! \mathbb{R}^{N_Y^{\ell}\times D^{\ell}}$ respectively, $N_X^{\ell}$ and $N_Y^{\ell}$ are the sizes of point sets, where we add the subscripts $X$, $Y$ to distinguish the source and target. By analyzing $\mathbf{F}_{X}^{\ell}$ and $\mathbf{F}_{Y}^{\ell}$ together using the Transformers $\eta_1^{\ell}$, $\eta_2^{\ell}$, the co-contextual information is collected as,
\begin{equation} 
\left\{ {\begin{aligned}
    \Phi_{\mathbf{X}^{\ell}}= \eta_1^{\ell}(\mathbf{F}_{X}^{\ell}, \mathbf{F}_{Y}^{\ell}) \\
    \Phi_{\mathbf{Y}^{\ell}}= \eta_2^{\ell}(\mathbf{F}_{Y}^{\ell}, \mathbf{F}_{X}^{\ell})
\end{aligned}} \right., %
\end{equation} 
where $\eta_1^{\ell}\!:\! \mathbb{R}^{N_X^{\ell}\times D^{\ell}}\! \times\! \mathbb{R}^{N_Y^{\ell}\times D^{\ell}}\!\to\! \mathbb{R}^{N_X^{\ell}\times D^{\ell+1}}$, and $\eta_2^{\ell}: \mathbb{R}^{N_Y^{\ell}\times D^{\ell}} \times \mathbb{R}^{N_X^{\ell}\times D^{\ell}} \to \mathbb{R}^{N_Y^{\ell}\times D^{\ell+1}}$. Then we select the feature subsets corresponding to the $\mathbf{X}^{\ell+1}$, $\mathbf{Y}^{\ell+1}$, notated as $\Phi_{\mathbf{X}^{\ell+1}} \in \mathbb{R}^{N_X^{\ell+1}\times D^{\ell+1}}$, $\Phi_{\mathbf{Y}^{\ell+1}} \in \mathbb{R}^{N_Y^{\ell+1}\times D^{\ell+1}}$, $N_X^{\ell+1}$ and $N_Y^{\ell+1}$ are point sizes of the next encoder layer.

\noindent\textbf{Different scale features extraction.}
In the $\ell$-th encoder layer, after capturing the results of both convolution branch and Transformer branch, we integrate them for the features of the next encoder layer, %
\begin{equation}
\left\{ {\begin{aligned}
    \mathbf{F}_{X}^{\ell+1}=\Psi_{\mathbf{X}^{\ell+1}} +\Phi_{\mathbf{X}^{\ell+1}} \\
    \mathbf{F}_{Y}^{\ell+1}=\Psi_{\mathbf{Y}^{\ell+1}} +\Phi_{\mathbf{Y}^{\ell+1}}
\end{aligned}} \right. .
\end{equation}

Then, by several encoder layers, different scale features are achieved. Notably, in each encoder layer, the information from both the current point cloud and the paired point cloud are mixed together. Compared with DCP \cite{wang_dcp_iccv_2019} and PRNet \cite{wang_prnet_nips_2019}, our method facilitates the matchability of correspondences further by exploiting and capturing co-contextual information from multiple scale features.

\subsection{Dynamic Fusion}
\label{sec::matching}
In the encoder network mentioned above, the original point set has been down-sampled layer-by-layer and the learned features become sparse corresponding to those down-sampled points with enlarged receptive fields. However, these different scale features always face a struggle between robustness and discriminativeness due to the unpredictable interference. 
We observe an intuitive rule, \ie the clean scale features are usually consistent in describing similar local geometries, while the contaminated ones are usually diverse due to the irregular disturbers. 
Then, we dynamically and adaptively fuse these obtained different scale features depending on the consistency of their distributions rather than a fixed trained decoder network. Essentially, these clean scale features cluster naturally, and we concentrate on them more and perform larger aggregation weights on them during our dynamic fusion process. In this way, the final obtained descriptor is guided to approach to the consistent and clean scale features while keeping away from the disturbed ones.

Specifically, we firstly propagate the obtained sparse point features to the dense point features corresponding to the original input point cloud, and then fuse these dense different scale features through our designed dynamic fusion module. 
Note that the source and target point clouds are handled in the same manner in this module, we take the source point cloud as an example for a clear explanation.

\noindent\textbf{Up-sampling for dense point features.} 
To achieve dense point features, we adopt a propagation strategy with distance-based interpolation here. Specifically, given the $\ell$-layer point features $\mathbf{F}^{\ell}\in\mathbb{R}^{N^{\ell}\times D^{\ell}}$ learned in the encoder network as mentioned above, we interpolate them to the coordinates of the original input point cloud $\mathbf{X}\in\mathbb{R}^{N\times3}$, where $N$ is the size of the original input point cloud. Notably, among the many choices for interpolation, we use the inverse distance weighted average based on K-nearest neighbors in our implementation, which is given as,
\begin{equation}
    \mathbf{f}_{\mathbf{x}_i}^{\ell\to \textsf{d}}=\frac{\sum_{k=1}^K w_k\mathbf{f}_{\mathbf{x}_k^\ell}}{\sum_{k=1}^{K}w_k},\quad  \mathbf{f}_{\mathbf{x}_k^\ell} \in \mathbf{F}^{\ell},
\label{eq:fp}
\end{equation}
where $w_k=1/{d(\mathbf{x}_i,\mathbf{x}_k^\ell)^{p}}$, $d(\cdot,\cdot)$ indicates the Euclidean distance, $p$ is the exponent, $\mathbf{x}_i\in\mathbf{X}$ is the current point and the $\mathbf{x}_k^\ell \in\mathbf{X}^{\ell}$ indicates the $k$-th neighbor point in the K-nearest neighbors of $\mathbf{x}_i$. %
Then, the \underline{d}ense feature of layer $\ell$ is returned as $\mathbf{F}_\textsf{d}^{\ell}=[\mathbf{f}_{\mathbf{x}_i}^{\ell\to \textsf{d}}]\in \mathbb{R}^{N\times D^{\ell}}$.
By the above up-sampling operation, all these different scale features have been unified to the same density $N$, \ie each original 3D point has same number of different scale features. Notably, in our implementation, the feature dimensions of different encoder layers are different. Then, we further unify these dense different scale features to a canonical dimension $v$ by applying MLP networks, \ie $\G^{\ell}=\textbf{MLP}_{\ell}(\mathbf{F}_\textsf{d}^{\ell})$,
where $\textbf{MLP}_{\ell}$ is the $\ell$-th MLP network. In this way, the dense feature $\G^{\ell}\in \mathbb{R}^{N\times v}$ capturing the $\ell$-th receptive field information is achieved. This up-sampling operation is performed for all encoder layers.

\noindent\textbf{Dynamic fusion of dense different scale features.} 
In this section, we fuse these different scale features for each 3D point. Formally, given the source and the target point clouds, different deformations and noises exist. Note the struggle between the robustness and discriminativeness of the single scale feature,
the strategy of joining different scale features has been proposed.
DGCNN \cite{wang_dgcnn_tog_2019} and PointNet++ \cite{charles_pointnet2_nips_2017} concatenate these different scale features, while KPConv \cite{thomas_kpconv_iccv_19} and D3Feat \cite{D3Feat_bai_cvpr_2020} apply a ResNet structure to add them together. However, these straightforward operations still cannot handle the disturbers because all different scale features are handled equally, where the contaminated ones are also embedded. Then the obtained final feature may play a counterproductive role in the matching task.
\figref{fig:ill} shows this problem intuitively. 
In response, we propose a dynamic fusion module here, which analyzes the distribution of these different scale features to recognize and pay more attention to the consistent ones during our dynamical fusion process. Because these consistent scale features are more likely to be the clean ones, our final fused descriptor achieves robustness and discriminativeness by mainly utilizing multiple clean scale feature.

As mentioned above, we denote the dense different scale feature as $\G^\ell = [\g^\ell] \in \mathbb{R}^{N \times v}$, $\g^\ell\in\mathbb{R}^{1\times v}$ is the feature of one point, $\ell=[1,2,...,L]$, $L$ is the number of scales. We take the different scale features of one point as an example. To fuse them, we aim at solving the corresponding dynamic coefficients $c^1,c^2,...,c^L$ to perform weighted average, which are determined dynamically in iteration in our EDFNet. Here, we take the $r$-th iteration procedure as an example to introduce our fusion algorithm clearly. Given the parameters $b_r^1,b_r^2,...,b_r^L$, let $c_r^1,c_r^2,...,c_r^L=softmax(b_r^1,b_r^2,...,b_r^L)$, then,
\begin{equation}
    \mathbf{s}_r=\sum\nolimits_{\ell=1}^{L}c_r^\ell \g^\ell.
\end{equation}
Next, we update the parameters according to the similarity between $\mathbf{s}_r$ and each scale feature, \ie
\begin{equation}
    b_{r+1}^\ell=b_r^{\ell}+\mathbf{s}_r{\g^\ell}^\mathrm{T}, \quad \ell=[1,2,...,L].
\end{equation}
Note that the coefficients of consistent scale features will increase as the iteration progresses due to the large similarity. After $T$ iterations, we get the final point feature descriptor as $\g^*=\sum_{\ell=1}^{L}c_T^\ell \g^\ell$. The whole iteration likes a clustering process to recognize the consistent ones with large coefficients, which will contribute more during the fusion. The complete fusion algorithm can be seen in \algref{alg:alg}.

Due to our dynamic fusion, the final obtained features approach to the consistent and clean scale features and keep away from the contaminated ones. Specifically, for the point far away from the disturbers, all scale features are consistent in describing a similar geometry structure. For the point near the disturbers, all scale features are disturbed. In these two rare cases, our dynamic fusion degenerates to equally processing like addition or concatenation, which is the bottleneck of our method for future study. However, in most cases, where disturbers exist in some larger scale features, the final fused feature is determined by the clean and consistent scale features more. Thus, under the premise of dodging the interference, we achieve the discriminative descriptor by jointly as many consistent and clean scale features as possible. 
Note that our dynamic fusion module is deep learning-free without any learning parameters. Thus, it determines the final fused feature descriptor totally depending on the different scale features themselves rather than a customized trained decoder network. This operation reiterates the generalization ability further when it is transferred to different datasets in practical applications.

\begin{algorithm}[!h]
	\KwIn{Number of scales $L$, features at different scales $\g^\ell$, $\ell=[1,2,...,L]$, iteration number $T$.}
	\KwOut{Fused point feature descriptor $\g^{*}$.}  
	\BlankLine
	
	Initialize parameters $b_0^1,b_0^2,...,b_0^L=0$.
	
	\While{$r \leqslant T-1$}{
		$c_r^1,c_r^2,...,c_r^L=softmax(b_r^1,b_r^2,...,b_r^L)$. \\ \qquad \textcolor{gray}{//dynamic fusion coefficients}
		
		$\mathbf{s}_r=\sum_{\ell=1}^{L}c_r^\ell \g^\ell$.

        $b_{r+1}^\ell=b_r^{\ell}+\mathbf{s}_r{\g^\ell}^\mathrm{T}$. \textcolor{gray}{//$\ell=[1,2,...,L]$}
	} 
	
	$c_T^1,c_T^2,...,c_T^L=softmax(b_T^1,b_T^2,...,b_T^L)$.
	
	$\g^*=\sum_{\ell=1}^{L}c_T^\ell \g^\ell$.

    \caption{Dynamic fusion algorithm.}
    \label{alg:alg}
\end{algorithm}

\subsection{Loss Function}\label{sec::loss}
In our method, the metric learning with the contrastive loss is used to train our EDFNet. Formally, given the source and target point clouds $\X$ and $\Y$, and a correspondence set $\Omega=\{(\mathbf{x}_i, \mathbf{y}_i)|i=1,...,n\}$, where the size of this set is $n$ and $(\mathbf{x}_i, \mathbf{y}_i)$ indicates the $i$-th correspondence, we notate the final learned descriptors of this $i$-th pair of points as $\bm{\gamma}_{\mathbf{x}_i}^{*}$ and $\bm{\gamma}_{\mathbf{y}_i}^{*}$ respectively. We use the Euclidean distance to measure the disparity between two descriptors. Then, the distance between the descriptors of a positive pair of points is defined as $d_i^\text{pos}=\|\bm{\gamma}_{\mathbf{x}_i}^{*}-\bm{\gamma}_{\mathbf{y}_i}^{*}\|_2$.
The distance between the descriptors of a negative pair of points is defined as
$d_i^\text{neg}=\text{min}( \|\bm{\gamma}_{\mathbf{x}_i}^{*}-\bm{\gamma}_{\mathbf{y}_j}^{*}\|_2) \ \text{s.t.}\ \|\mathbf{y}_j - \mathbf{y}_i \|_2 > \varepsilon$,
where $\varepsilon$ is the safe radius, $\mathbf{y}_j$ is the hardest negative sample that lies outside the safe radius of the correct corresponding point $\mathbf{y}_i$. Then, our final loss function is defined as,
\begin{equation}
    \mathcal{L} \!=\! \frac{1}{n} \sum_i \left[ \text{max}\left(0, d_i^\text{pos} \!-\!M^\text{pos}\right) + \text{max}\left(0, M^\text{neg}\!-\!d_i^\text{neg}\right) \right],
    \label{equ:des_L}
\end{equation}
where $M^\text{pos}$ indicates the margin for positive pairs and $M^\text{neg}$ indicates the margin for negative pairs.

\subsection{Implementation Details} \label{implementaton} %
During the training, we apply data augmentation to each point cloud including the Gaussian jitter with mean 0, standard deviation 0.005, the random scaling $\in [0.9, 1.1]$, and the random rotation angle $\in [0^\circ, 360^\circ]$ along an arbitrary axis.
We set the number of scales $L=5$.
Then, we use grid down-sampling to reduce the size of the input point cloud layer-by-layer.
The numbers of feature channels are set to $[64, 128, 256, 512, 1024]$ in five encoder layers, respectively. In up-sampling procedure, the parameter $K$ in K-nearest neighbor is set to 24. And all point features are finally uniformed to 32 dimension. 
The positive margin $M^\text{pos}=0.1$ and negative margin $M^\text{neg}=1.4$ in our contrastive loss. The size of the correspondence set $n=64$.
The iteration number $T=5$.
Our EDFNet is implemented in Pytorch 1.7.0 and trained on RTX 3090. We use the SGD optimizer with momentum parameter 0.98 and weight decay parameter $1e^{-6}$ to optimize the network with the initial learning rate of 0.1 for 150 epochs. 
The learning rate scheduler adopts exponentially decayed with gamma parameter $0.97$. %

\section{Experiments and Evaluation}\label{sec:experiment}
In this section, we report the accuracy and robustness of our proposed EDFNet regarding 3D point cloud matching. To this end, we evaluate our method on several benchmark datasets including the typical indoor dataset, 3DMatch \cite{3dmatch_zeng_cvpr_2017}, and the outdoor dataset, KITTI \cite{kitti_andreas_ijrr_2013}. Besides, to verify the robustness and generalization ability of our method, we perform the test on both rigid and non-rigid point cloud data further.

\subsection{Indoor Dataset: 3DMatch} \label{sec::3dmatch}
The training and testing sets of the 3DMatch dataset, and the ground truth transformation have been provided officially \cite{3dmatch_zeng_cvpr_2017}. The ground truth correspondences are constructed according to the nearest neighbor principle. Specifically, if the distance between the transformed point, achieved by the ground truth transformation, and the nearest point is less than a pre-defined threshold (5cm), this correspondence is confirmed as a ground truth correspondence. Following \cite{D3Feat_bai_cvpr_2020}, during the training, all the point cloud fragment pairs with at least 30\% overlap are selected from the official training set. The test set contains 8 scenes with partially overlapped point cloud fragments. All fragments have been down-sampled with 2.5cm voxelsize.

\noindent\textbf{Evaluation metrics.} Following \cite{D3Feat_bai_cvpr_2020,choy_fcgf_iccv_2019}, three evaluation metrics are used including \texttt{Inlier Ratio}, \texttt{Feature Matching Recall}, and \texttt{Registration Recall}.
Specifically, given two point clouds $\mathbf{X}$ and $\mathbf{Y}$, a feature learning network maps the input points to feature descriptors. Then the index set of predicted correspondences is notated as $\pi=\{(i,j)\}$, where $\mathbf{x}_i \in \mathbf{X}$ and $\mathbf{y}_j \in \mathbf{Y}$ are predicted correspondences obtained by nearest neighbor searching in the feature space. 

\noindent$\bullet$~\texttt{Inlier Ratio}. This metric reports the mean percentage of successfully identified inliers. Given a set of point cloud pairs $M$, the inlier ratio $\mathcal{I}$ of one pair $m$ is defined as,
\begin{equation}\vspace{-0.2cm}
    \mathcal{I}=\frac{1}{|\pi|}\sum_{(i,j)\in \pi} \bm{1}(\|T_m(\mathbf{x}_i) - \mathbf{y}_j\|<\tau_1),
\end{equation}
where $T_m(\cdot)$ is the ground truth transformation of the point cloud pair $m$. $\bm{1}(\cdot)$ is a binary function. It equals to 1 if the input is true, otherwise it equals to 0. $\tau_1$ is the inlier distance threshold, which is set to $\tau_1=10cm$ following \cite{3dmatch_zeng_cvpr_2017}. Finally, we report the mean of inlier rations of all pairs of point clouds as the evaluation result.

\noindent$\bullet$~\texttt{Feature Matching Recall}. This metric is defined as the percentage of successful alignment whose inlier ratio is above a threshold to measures the matching quality of pairwise registration. Then, the feature matching recall is defined as,
\begin{equation}\vspace{-0.2cm}
    \eta_\text{fmr}=\frac{1}{|M|}\sum_{M}\bm{1}(\mathcal{I} > \tau_2 ),
\end{equation}
where $\tau_2$ is the inlier ratio threshold of the point cloud pair. And the point cloud pairs with more than $\tau_2=5\%$ inliers will be counted as one successfully matched pair. 

\noindent$\bullet$~\texttt{Registration Recall}. This metric is defined as the percentage of successful alignment whose transformation estimation error is below a threshold. At first, we use a robust local registration algorithm, RANSAC with 50000 max iterations, to estimate the rigid transformation between the two input point clouds. Then, the RMSE distance of the ground truth correspondences under the estimated transformation matrix is calculated. The ground truth correspondence set for the point cloud pair $m$ is given as $\Omega^* = \{\mathbf{x}^* \in \mathbf{X}, \mathbf{y}^*\in \mathbf{Y}\}$.
Then, the RMSE distance notated as $\text{RMSE}_\text{dis}$ of this point cloud pair $m$ is defined as, 
\begin{equation}\vspace{-0.2cm}
    \text{RMSE}_\text{dis} = \sqrt{\frac{1}{|\Omega^*|}\sum_{(\mathbf{x}^*, \mathbf{y}^*)\in\Omega^*}\|\hat{T}_m(\mathbf{x}^*)-\mathbf{y}^*\|^2},
\end{equation}
where $\hat{T}_{m}(\cdot)$ is the estimated transformation of the point cloud pair $m$ returned by RANSAC. Finally, the registration recall is defined as,
\begin{equation}
    \eta_\text{rr}=\frac{1}{|M|}\sum_{M}\bm{1}\left(\text{RMSE}_\text{dis}<\tau_3
    \right),
\end{equation}
where $\tau_3$ is set to 0.2 following \cite{D3Feat_bai_cvpr_2020}.

\noindent\textbf{Evaluation points sampling.}
Following \cite{D3Feat_bai_cvpr_2020}, during the test, we select 5000 points to conduct the evaluation rather than taking all points. Here, two strategies of points selection are employed, \ie randomly sampled points (\texttt{rand}) and keypoints predicted by a detector (\texttt{pred}). For the first one, we randomly select a specified number of points from the pairs of the point cloud fragments for evaluation. For the second one, we use the detector proposed in \cite{D3Feat_bai_cvpr_2020}, which solves a score to indicate the matchability of each point. We select a specified number of points with higher scores for our evaluation. Notably, in our implementation of the \texttt{pred} setting, we add the detector after our original EDFNet, and train the network using the equal weighted loss functions of ours in \eqref{equ:des_L} and the detector loss function provided in \cite{D3Feat_bai_cvpr_2020}.
Please refer to \cite{D3Feat_bai_cvpr_2020} for more details about the detector.

\begin{table}[!h]
	\renewcommand\arraystretch{1.0}
	\vspace{-0.1cm}
	\caption{Feature matching recall at $\tau_1=10\text{cm}$, $\tau_2=5\%$ on the 3DMatch dataset. \textbf{AVG} and \textbf{STD} indicate the average feature matching recall and its standard deviation. \textbf{Dim.} indicates the dimension of the final feature. Time consumption of per feature is tested on Intel Core i7-11700K @ 3.60GHz and Nvidia RTX3090.}
	\vspace{-0.3cm}
	\begin{center}
	    \resizebox{0.85\linewidth}{!}{
			\begin{tabular}{l|cccc}
				\toprule
			    \textbf{Methods}&\textbf{AVG}(\%)&\textbf{STD} &\textbf{Dim.} & Times(ms)\\
				\midrule
				{Spin \cite{andrew_usi_pami_1999}}       &22.7  &11.4  &153  &0.093\\
				{FPFH \cite{rusu_fpfh_icra_2009}}        &35.9  &13.4  &33   &0.022\\
				{SHOT \cite{SHOT_Tombari_2010_ECCV}}     &23.8  &10.9  &352  &0.201\\
				{USC \cite{USC_Tombari_10}}              &40.0  &12.5  &1980 &2.483\\
				{PointNet \cite{charles_pointnet_cvpr_2017}}&47.1 &12.7&256  &0.117\\
				{3DMatch \cite{3dmatch_zeng_cvpr_2017}}  &59.6  &8.8   &512  &2.151\\
				{CGF \cite{CGF_marc_iccv_2017}}          &58.2  &14.2  &32   &0.981\\
				{PPFNet \cite{PPFNet_Deng_2018_CVPR}}    &62.3  &10.8  &64   &1.510\\
				{PPF-FoldNet \cite{PPFFolding_Deng_2018_ECCV}}  &71.8 &10.5 &512 &0.531\\
				{DirectReg \cite{deng_DirectReg_cvpr_2019}}&74.6&9.4   &512  &0.531\\
				{CapsuleNet \cite{zhao_3dcapsule_cvpr_2019}}&80.7&6.2  &512  &0.808\\
				{PerferctMatch \cite{perfect_zan_cvpr_2019}}  &94.7  &2.7 &32 &3.678\\
				{FCGF \cite{choy_fcgf_iccv_2019}}        &95.2  &2.9   &32   &0.013\\
				{D3Feat(rand) \cite{D3Feat_bai_cvpr_2020}}&95.3 &2.7   &32    &\textbf{0.005}\\
				{D3Feat(pred) \cite{D3Feat_bai_cvpr_2020}}&95.8 &2.9   &32    &\textbf{0.005}\\
				{LMVD \cite{li_mvd_cvpr_2020}} &  97.5 & 2.8 & 32 & - \\
				{SpinNet \cite{Ao_spinnet_cvpr_2021}} &\textbf{97.6} & 1.9 & 32 & 4.216 \\
				DGR \cite{choy_dgr_cvpr_2020}      &97.1 &2.7 &32 &0.842 \\
				\cdashline{1-5}[2.2pt/1.2pt] 
				{EDFNet(rand) }                          &95.8  &2.7   &32    &0.016\\
				{EDFNet(pred) }                 &{97.5}  &2.7   &32    &0.016\\
				\bottomrule
			\end{tabular}}
	        \label{tab:3dmatch:fmr}
	\end{center}
	\vspace{-0.5cm}
\end{table}

\noindent\textbf{Evaluation.} In this part, we compare our EDFNet with the \textit{SOTA} methods on the 3DMatch dataset. In \tabref{tab:3dmatch:fmr}, we report the average feature matching recall result and its standard deviation indicated by \textbf{AVG} and \textbf{STD} respectively. The results of the handcrafted features, such as Spin \cite{andrew_usi_pami_1999}, SHOT \cite{SHOT_Tombari_2010_ECCV}, usually fail in such a challenging dataset. The learning-based methods perform better than these handcrafted methods. Our proposed EDFNet achieves the second-top performance among all methods, whose results are the same as LMVD \cite{li_mvd_cvpr_2020}. 
SpinNet \cite{Ao_spinnet_cvpr_2021} achieves a slightly better result than ours (0.1\%) due to the customized rotation invariant operation based on the rigid transformation assumption. This makes its obtained descriptor more specific. And SpinNet achieves more accurate results in this evaluation on the 3DMatch dataset because a rigid transformation exists in the input point cloud pair. By contrast, ours is devoted to learning a matching task-specific descriptor based on the local geometry structure of the input point clouds. Thus, our EDFNet can be extended to handle the non-rigid point cloud data as presented in \secref{sec:generalization}, which brings more general applications.

Besides, we also provide the time consumption per feature. D3Feat presents the best time-efficiency, and ours is slightly slower than D3Feat and FCGF due to the multiple Transformer modules and the iteration in the dynamic fusion module. Note that DGR \cite{choy_dgr_cvpr_2020} performs a further optimization on the initial matches, which are determined based on the FCGF descriptor. DGR learns a weight to indicate the reliability of each match. Here, after normalizing the weight coefficients, we select the matches whose weight $\iota>0.5$ to conduct the evaluation. In \tabref{tab:3dmatch:fmr}, we can see that DGR presents a better feature matching recall result than FCGF after this further optimization at the cost of discarding many matches whose weight $\iota \le 0.5$. Nevertheless, our proposed EDFNet still outperforms DGR.

\begin{figure}[!h]
    \vspace{-\baselineskip}
	\centerline{\includegraphics[width=0.88\linewidth]{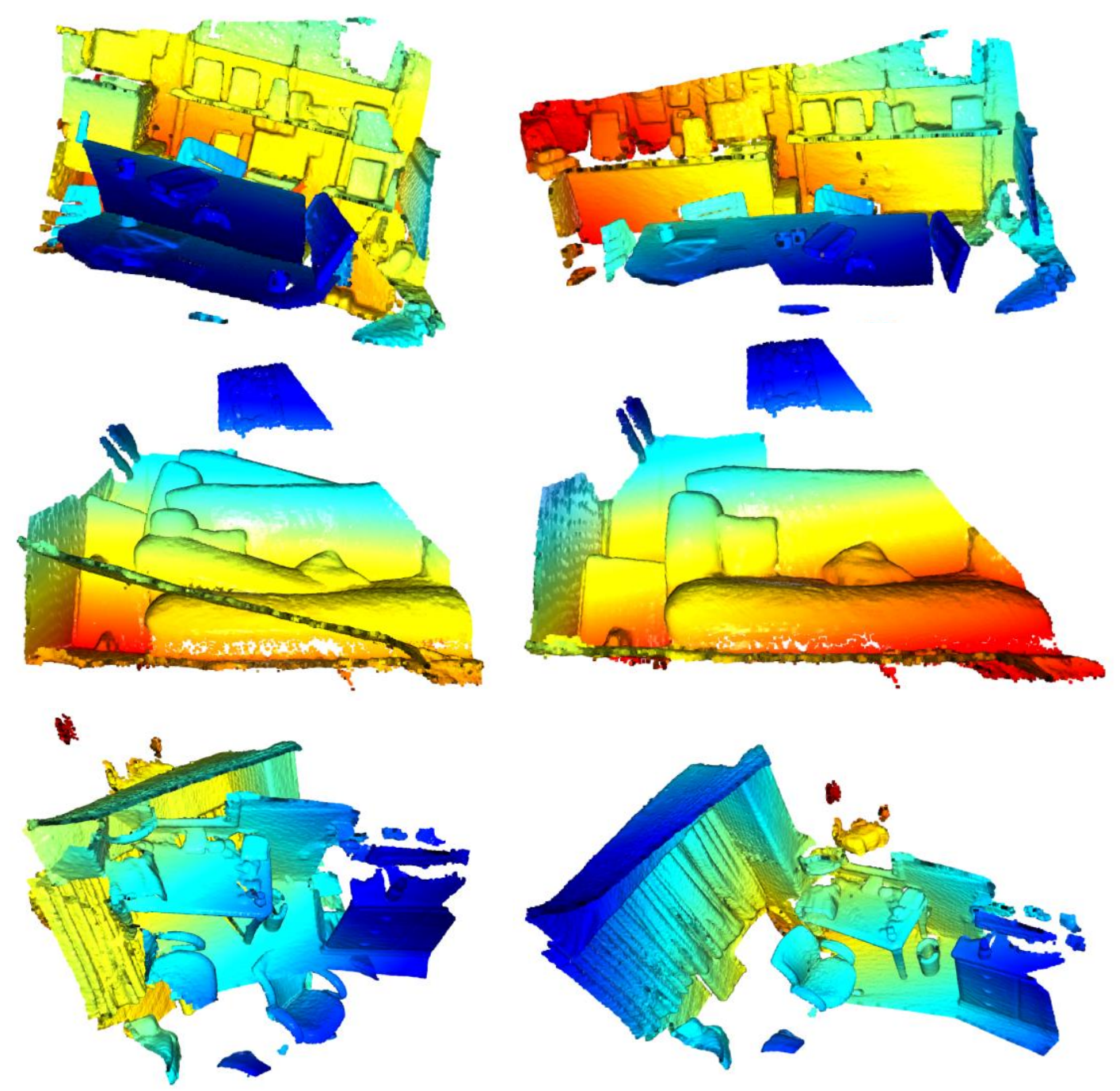}}
	\vspace{-0.2cm}
	\caption{Visualization of registration results on the 3DMatch dataset. The left column indicates two partial input point clouds with significant rigid transformation. The right column indicates the accurate registration result achieved by performing the RANSAC on the point correspondences constructed based on our learned descriptors.}
	\label{fig:3dmatch}
	\vspace{-0.5cm}
\end{figure}

\begin{figure}[!h]
	\centerline{\includegraphics[width=\linewidth]{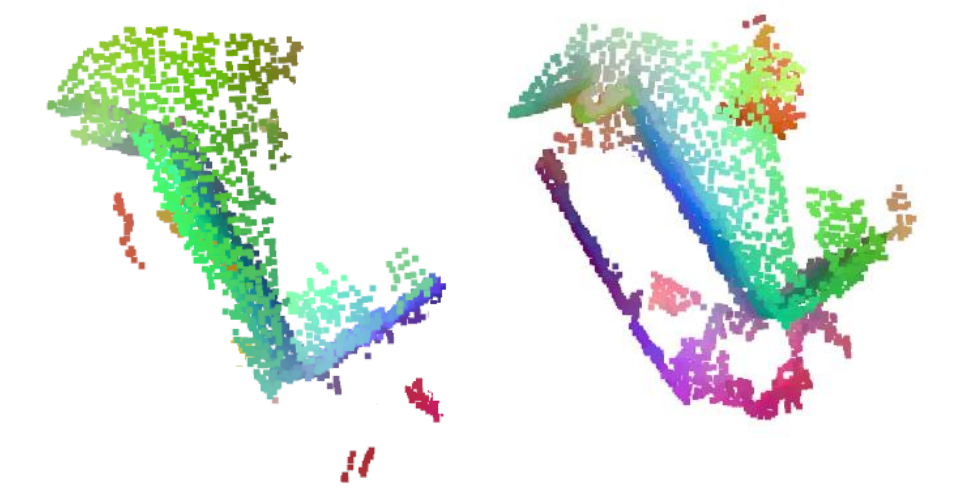}}
	\vspace{-0.2cm}
	\caption{Visualization of the distribution of our learned feature descriptor. Left is the source point cloud and right is the target point cloud. Each high-dimensional feature is reduced to 3-dimension by the T-SNE algorithm. Then, we normalize them to the color space to represent the color of each point. We can find that the correspondences tend to be marked with the same color. In other words, consistent feature descriptors are achieved for correspondences, which validates the effectiveness of our EDFNet in the point cloud matching task.} 
	\label{fig:3dmatch_feat}
	\vspace{-0.5cm}
\end{figure}

According to the matching results based on the learned point feature descriptor, we accurately register the two input point clouds. To validate our method intuitively, we visualize some registration results in \figref{fig:3dmatch}, from which we can find that our method achieves accurate registration.
Besides, we also present the qualitative results of the distribution of the learned task-specific descriptors. Specifically, we reduce the dimension of the descriptors to 3-dimension by T-SNE algorithm to represent the RGB in color space. Then, we show the pair of point clouds where each point is marked with the corresponding color. From \figref{fig:3dmatch_feat}, we find that the point correspondences tend to be marked with the same color under the interference, which validates the effectiveness of our learned descriptor for the matching task.

\subsection{Outdoor Dataset: KITTI} \label{sec:kitti}
In this section, we evaluate our method on the outdoor dataset, KITTI, which consists of 11 sequences of outdoor driving scenarios. The sequences 0-5 are used for training, 6-7 for validation, and 8-10 for testing following \cite{choy_fcgf_iccv_2019}. Since the GPS ground truth is noisy, we firstly use the standard ICP \cite{besl_icp_pami_1992} to refine the ground truth transformation. The ground truth correspondences are determined as the 3DMatch, where the threshold is set to 5cm. We select LiDAR scan pairs with at least 10m intervals to construct our input point cloud pairs. All point clouds are voxel down-sampled with 30cm voxelsize.

\noindent\textbf{Evaluation metrics.} Following DGR \cite{choy_dgr_cvpr_2020}, we evaluate the estimated transformation matrices by \texttt{Translation Error} \texttt{(TE)} and \texttt{Rotation Error} \texttt{(RE)}. The rotation error is calculated as,
\begin{equation}\vspace{-0.2cm}
    \text{RE} = \text{arccos}\left(\frac{\text{Tr}(\mathbf{R}_\text{pre}^{\mathrm{T}} \mathbf{R}_\text{gt})-1}{2}\right),
\end{equation}
where $\mathbf{R}_\text{gt}$ and $\mathbf{R}_\text{pre}$ indicate the ground truth and predicted rotation matrix. $\text{Tr}(\cdot)$ solves the trace of the input matrix. The translation error is calculated by,
\begin{equation}\vspace{-0.2cm}
    \text{TE} = \|\mathbf{t}_\text{pre} - \mathbf{t}_\text{gt}\|_2^2,
\end{equation}
where $\mathbf{t}_\text{pre}$ and $\mathbf{t}_\text{gt}$ indicate the predicted and ground truth translation vector, respectively. Besides RE and TE, we also report the recall results, which indicate the percentage of successful pairwise registration. The successful registration is confirmed if TE $\le 0.6m$ and RE $\le 5^\circ$ following \cite{choy_dgr_cvpr_2020}. 
Average TE and RE reported in \tabref{tab:kitti:reg} are computed only on these successfully registered pairs since the poses returned by failed registrations can be drastically different from the ground truth, making the error metrics unreliable.

\begin{table}[!h]
	\renewcommand\arraystretch{1.0}
	\caption{Registration on the KITTI dataset. \textbf{AVG} indicates the average result and \textbf{STD} indicates the standard deviation.}
	\vspace{-0.4cm}
	\begin{center}
	    \resizebox{0.85\linewidth}{!}{
			\begin{tabular}{l|cc|cc|c}
				\toprule
			    \multirow{2}*{{\textbf{Methods}}}&\multicolumn{2}{c|}{\textbf{TE}(cm)}&\multicolumn{2}{c|}{\textbf{RE}($^\circ$)}&\multirow{2}*{\textbf{Recall}(\%)} \\
			    \cline{2-3}\cline{4-5}
			    ~&\multicolumn{1}{c}{\textbf{AVG}}&\multicolumn{1}{c|}{\textbf{STD}}&\multicolumn{1}{c}{\textbf{AVG}}&\multicolumn{1}{c|}{\textbf{STD}}&~\\
				\midrule
				{FGR \cite{zhou_fgr_eccv_2016}}                    &42.8  &19.7  &1.89  &1.11  & 2.74  \\  
				{3DFeat-Net \cite{yew_3dfeatnet_eccv_2018}}        &21.4  &17.2  &0.44  &0.36  & 45.72 \\
				{FCGF \cite{choy_fcgf_iccv_2019}}                  &10.5  &1.37  &0.30  &0.29  & 97.34 \\
				{D3Feat \cite{D3Feat_bai_cvpr_2020}}               &6.54  &1.06  &0.33  &0.23  & 97.81 \\
				{SpinNet \cite{Ao_spinnet_cvpr_2021}}        &9.13  &1.42 &0.73 &0.41 & 97.50\\
				{EDFNet}                                     &5.12  &1.01  &0.31  &0.20  & 98.14 \\
				\midrule
				{DGR \cite{choy_dgr_cvpr_2020}}                    &4.01  &0.82  &0.23  &0.14  & 97.94 \\
				{PointDSC \cite{PointDSC_Bai_2021_CVPR}}           &3.84  &0.83  &0.22  &{0.12}  & 98.70 \\
				{EDFNet+DGR}                                        &\textbf{3.82}  &{0.77}  &\textbf{0.17}  &0.14  & \textbf{98.92} \\
				\bottomrule
			\end{tabular}}
	        \label{tab:kitti:reg}
	\end{center}
	\vspace{-0.8cm}
\end{table}

\begin{figure}[!h]
	\centerline{\includegraphics[width=1.0\linewidth]{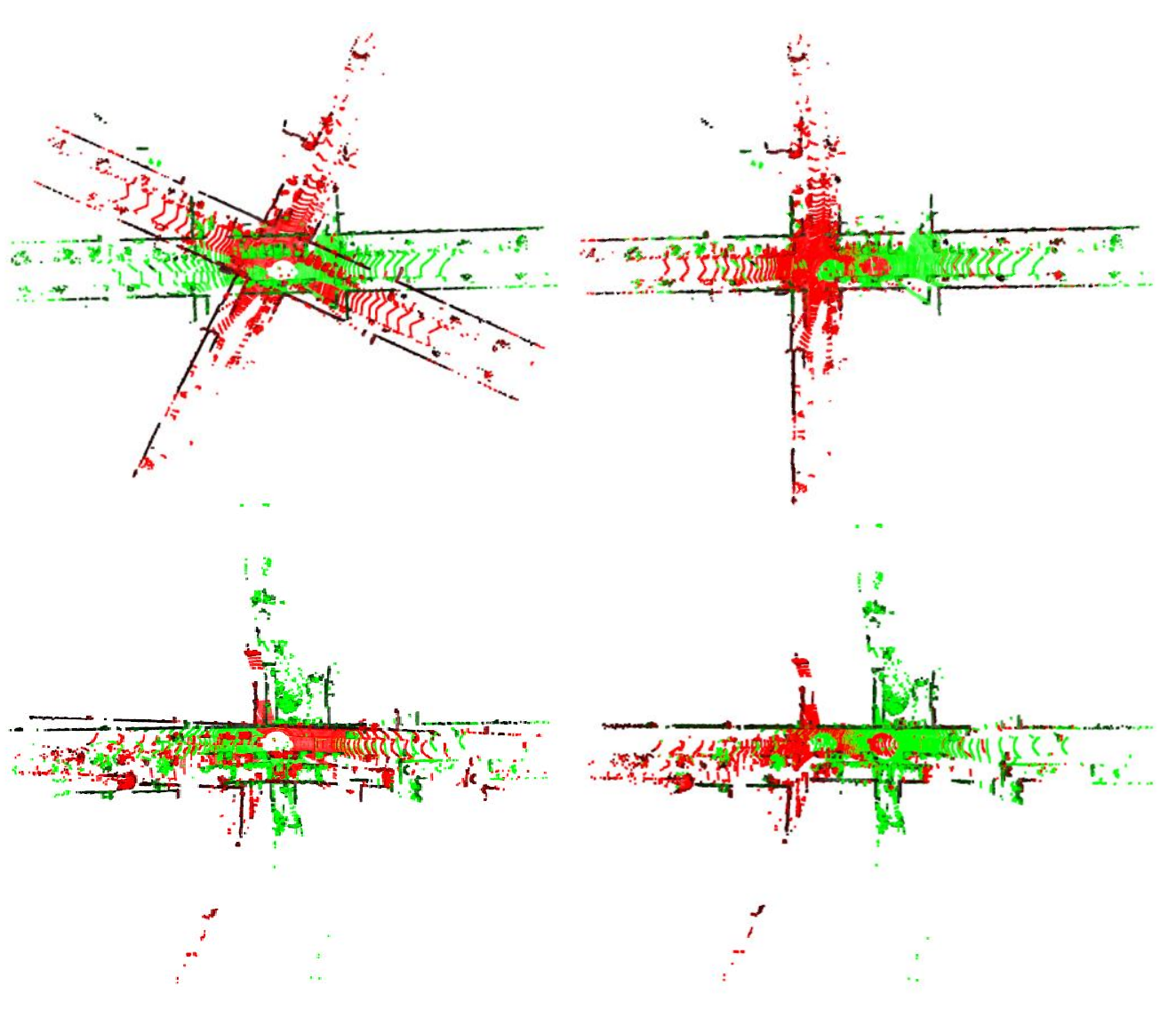}}\vspace{-0.2cm}
	\caption{Registration results on the KITTI dataset. The left is the input point cloud pair while the right is the registration result. The proposed descriptor used for the matching results in accurate registration performance.}
	\label{Fig:kitti}
	\vspace{-0.4cm}
\end{figure}

\noindent\textbf{Evaluation.}
We compare our EDFNet with FCGF, 3DFeat-Net, and D3Feat, which are committed to extracting advanced point feature descriptors. Besides, we also compare our method with \textit{SOTA} learning-based point cloud registration methods, including DGR \cite{choy_dgr_cvpr_2020} and PointDSC \cite{PointDSC_Bai_2021_CVPR}, which focus on selecting reliable correspondences from the initially constructed correspondences. Notably, DCP \cite{wang_dcp_iccv_2019} and PRNet \cite{wang_prnet_nips_2019} are divergent in this setting. We evaluate two versions of EDFNet. 
For the first version, the final transformation is obtained by performing RANSAC with 50,000 max iterations after initial nearest neighbor feature matching. The registration results of FGR, 3DFeat-Net, FCGF, SpinNet and D3Feat are also achieved by this operation. 
For the second version, we use the proposed EDFNet to extract point features instead of the original FCGF in DGR framework to estimate the final transformation. 
The registration results are provided in \tabref{tab:kitti:reg}, where EDFNet outperforms the \textit{SOTA} point feature descriptor learning methods in all metrics. But it underperforms the learning-based registration methods, DGR \cite{choy_dgr_cvpr_2020} and PointDSC \cite{PointDSC_Bai_2021_CVPR} since they optimize the matching result further after point feature learning.
In addition, the final registration performance is further improved when our method is integrated into the advanced learning-based point cloud registration framework DGR.
Moreover, to illustrate our method intuitively, we present some typical registration results of EDFNet + DGR in \figref{Fig:kitti}, where ours achieves successful registration.

\subsection{Generalization evaluation} \label{sec:generalization}
\noindent\textbf{Evaluation on rigid point cloud data.} To evaluate the robustness and generalization ability of our EDFNet, we use the model trained on the 3DMatch dataset to conduct the test on four outdoor laser scan datasets (Gazebo-Summer, Gazebo-Winter, Wood-Summer and Wood-Autumn) from the ETH dataset \cite{eth_pomerleau_ijrr_2012} following \cite{perfect_zan_cvpr_2019}. The evaluation metric is feature matching recall, and the experiments adopt the same settings as the previous evaluation on 3DMatch. 
The baselines involve two settings for the voxelsize, \ie 5cm and 6.25cm. For a fair and comprehensive comparison, we evaluate ours in these two settings.
The results of PerfectMatch with 6.25cm voxelsize and FCGF with 5cm voxelsize are provided. For D3Feat, we report the results on both 6.25cm and 5cm voxelsizes. 
As shown in \tabref{tab:eth}, under the 5cm voxelsize setting, our EDFNet presents better generalization ability than other baselines. Under the 6.25cm voxelsize setting, PerfectMatch achieves better results than our EDFNet on the Gazebo-Summer and Gazebo-Winter. However, for Wood-Summer and Wood-Autumn, ours outperforms PerfectMatch. And our EDFNet is better than PerfectMatch in terms of the final average results. We suspect that the reason is that there are more noise and outliers in the Wood sub-dataset than the Gazebo sub-dataset, which decreases the performance of PerfectMatch. And our EDFNet exhibits more robustness and presents more stable matching results. 
LMVD \cite{li_mvd_cvpr_2020} and SpinNet \cite{Ao_spinnet_cvpr_2021} do not report their voxelsize, while SpinNet obtains the most accurate results among all methods whether 5cm voxelsize or 6.25cm voxelsize. We suspect the reason is that SpinNet designs a learning-free proactive operation for rotation invariance, which helps improve the generalization ability significantly. Our learning-free module, dynamic fusion, is post-processing after different scale features learning, which is affected more heavily than SpinNet. Nevertheless, ours achieves the second-top performance among all methods on the \textbf{AVG} metric. DGR is conducted following the setting mentioned in \secref{sec::3dmatch}. We can find that it outperforms FCGF significantly by discarding many matches with low reliability weights. However, DGR still underperforms our proposed EDFNet. We suspect that the reason is its limited generalization ability.

\vspace{-0.5cm}

\begin{table}[!ht]
	\renewcommand\arraystretch{1.0}
	\caption{Feature matching recall comparison on the ETH dataset, where $\tau_1=10cm$, $\tau_2=5\%$. The model is trained on the 3DMatch dataset. Some results are copied from \cite{D3Feat_bai_cvpr_2020,Ao_spinnet_cvpr_2021}.}
	\vspace{-0.3cm}
	\begin{center}
	    \resizebox{0.9\linewidth}{!}{
			\begin{tabular}{l|c|cc|cc|c}
				\toprule
			    \multirow{2}*{{\textbf{Methods}}}&\multicolumn{1}{c|}{\textbf{voxel}}&\multicolumn{2}{c|}{\textbf{Gazebo}}&\multicolumn{2}{c|}{\textbf{Wood}}&\multirow{2}*{\textbf{AVG}(\%)} \\
			    \cline{3-4}\cline{5-6}
			    ~&\multicolumn{1}{c|}{\textbf{size} (cm)}&\multicolumn{1}{c}{Sum.}&\multicolumn{1}{c|}{Wint.}&\multicolumn{1}{c}{Sum.}&\multicolumn{1}{c|}{Aut.}&~\\
				\midrule
				{PerfectMatch \cite{perfect_zan_cvpr_2019}}   &6.25     &{91.3}  &{84.1}  &{67.8}  &{72.8}  &{79.0} \\
				{FCGF \cite{choy_fcgf_iccv_2019}}             &5.00     &22.8  &10.0  &17.8  &16.8  & 16.1 \\
				{D3Feat(rand) \cite{D3Feat_bai_cvpr_2020}}    &5.00     &45.7  &23.9  &13.0  &22.4  & 26.2 \\
				{D3Feat(pred) \cite{D3Feat_bai_cvpr_2020}}    &5.00     &78.9  &62.6  &45.2  &37.6  & 56.3 \\
				{D3Feat(pred) \cite{D3Feat_bai_cvpr_2020}}    &6.25     &85.9  &63.0  &49.6  &48.0  & 61.6 \\
				LMVD \cite{li_mvd_cvpr_2020} &-& 85.3 & 72.0 & 84.0 & 78.3 & 79.9\\
				SpinNet \cite{Ao_spinnet_cvpr_2021} &-& \textbf{92.9} & \textbf{91.7} & \textbf{92.2} & \textbf{94.4} & \textbf{92.8}\\
				DGR \cite{choy_dgr_cvpr_2020}     &5.00 &74.2 &66.7 &70.1 &64.6 &67.5\\
				\cdashline{1-7}[2.2pt/1.2pt]
				{EDFNet(rand)}                                 &5.00     &60.3  &52.7  &40.1  &38.4  & 49.3 \\
				{EDFNet(pred)}                                 &5.00     &80.4  &67.3  &72.0  &69.8  & 70.7 \\
				{EDFNet(pred)}                                 &6.25     &88.4  &82.7  &84.3  &79.1  & 80.5 \\
				\bottomrule
			\end{tabular}}
	        \label{tab:eth}
	\end{center}
	\vspace{-0.6cm}
\end{table}	

\noindent\textbf{Evaluation on non-rigid point cloud data.} 
To validate the robustness and generalization ability of our proposed method further, we conduct a challenging experiment on non-rigid point cloud data following \cite{zeng_corrnet3d_cvpr_2021}. Specifically, we adopted Surreal \cite{groueix_3dcoded_eccv_18} as the training dataset, consisting of 230K samples, which were randomly grouped into 115K training pairs. We conducted the test on the SHREC dataset \cite{donati_deepgfm_cvpr_20}, which has 430 pairs of non-rigid shapes. The test is conducted on the whole point cloud without point selection.

\begin{figure}[!h]
    \vspace{-\baselineskip}
	\centerline{\includegraphics[width=0.92\linewidth]{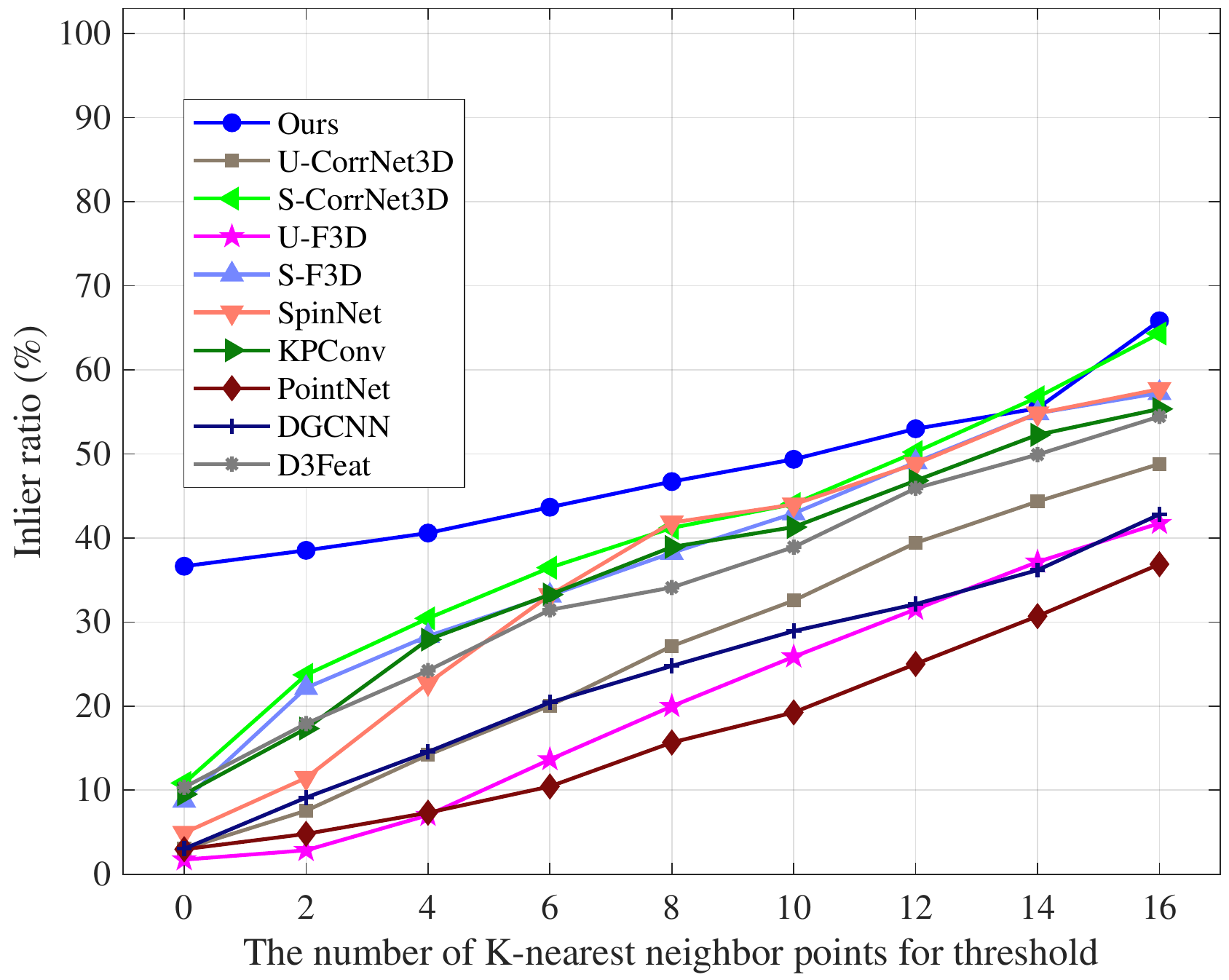}}
	\vspace{-0.2cm}
	\caption{Comparison of the inlier ratio results on non-rigid point clouds of Surreal\textbackslash SHREC. Note: $K=0$ means that the self-adaptive threshold is set to 0. And our proposed EDFNet achieves the comparable performance than the \textit{SOTA} supervised CorrNet3D.
	}
	\label{fig:nonrigid}
\end{figure}

We report the average \texttt{Inlier Ratio} here to evaluate the matching performance. Different from the version used before, where the threshold to confirm the correspondence inlier is set as a fixed distance value, we adopt a self-adaptive threshold as follows. To avoid confusion caused by the fixed distance based threshold $\tau_1$, we notate this self-adaptive threshold as $\kappa$, which is formulated based on the K-nearest neighbor principle. Specifically,
${\kappa _i} = \frac{1}{K} {\sum_{j = 1}^K {{{l}_{\mathbf{x}_{i},\mathbf{x}_j}}}}$, $\mathbf{x}_{j} \in \text{KNN} \left( \mathbf{x}_{i} \right)$, where ${l_{\mathbf{x}_{i},\mathbf{x}_{j}}}$ is the Euclidean distance between the points $\mathbf{x}_{i}$ and $\mathbf{x}_{j}$, $\text{KNN}(\cdot)$ represents the K-nearest neighbor. In other words, for the $i$-th point $\mathbf{x}_{i}$, $\kappa_i$ is computed as the mean distances between the point $\mathbf{x}_{i}$ and its $K$ nearest neighbor points.

\begin{figure}[!h]
    \vspace{-\baselineskip}
	\centerline{\includegraphics[width=\linewidth]{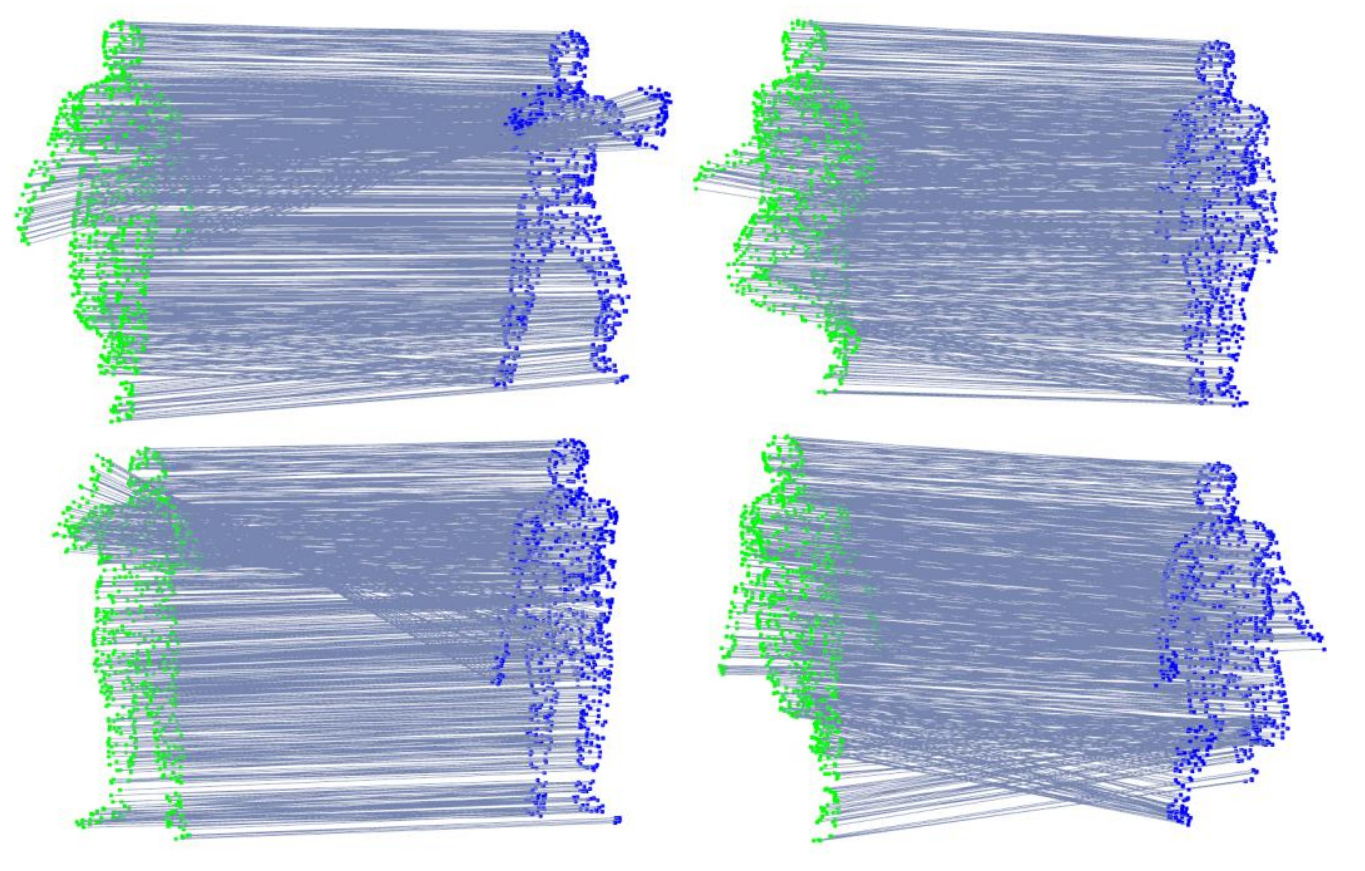}}
	\vspace{-0.2cm}
	\caption{Matching results of our method on non-rigid point clouds. Green indicates the source point cloud and blue indicates the target. Lines present the matching results, where our method achieves advanced performance.}
	\label{fig:nonrigid_vis}\vspace{-0.3cm}
\end{figure}

As shown in \figref{fig:nonrigid}, we report the results according to different $K$ value. During the evaluation, the nearest neighbor principle in feature space is applied to determine the matching after point feature learning. For PointNet \cite{charles_pointnet_cvpr_2017}, DGCNN \cite{wang_dgcnn_tog_2019}, the metric learning with contrastive loss function is used to train the network to learn the point features. Besides the point feature learning based methods, two versions of CorrNet3D \cite{zeng_corrnet3d_cvpr_2021}, \ie supervised version (notated as S-CorrNet3D) and unsupervised version (U-CorrNet3D) are also evaluated. 
U-F3D and S-F3D indicate unsupervised FlowNet3D \cite{liu_flownet3d_cvpr_2019} and supervised FlowNet3D respectively. Among all point feature learning based methods, ours achieves the best matching results. 
Notably, SpinNet is designed for the rigid point cloud, which underperforms ours significantly.
Comparing to the \textit{SOTA} point cloud matching method, SpinNet, besides the time efficiency as introduced in \secref{sec::3dmatch}, our EDFNet presents another advantage here. Since SpinNet is designed based on the assumption of rigid transformation, it fails and performs decreased performance when handling the non-rigid point cloud data. By contrast, our EDFNet also achieves reliable results because it is built by focusing on local geometry structure learning, which enables more broad applications.
Meanwhile, compared with the \textit{SOTA} method, S-CorrNet3D, ours also achieves comparable or even better performance, which validates the generalization ability of our method sufficiently. Meanwhile, for an intuitive understanding of our method, we provide some matching results on these non-rigid point cloud pairs in \figref{fig:nonrigid_vis}. Obviously, our method achieves advanced matching performance.

\noindent\textbf{Evaluation on non-rigid point cloud sequences data.} To demonstrate the matching effectiveness and temporal consistency of our proposed method, we test our EDFNet on a dynamic point cloud sequences dataset. We adopt the 8iVFB dataset \cite{eon_8ivfb_2017} here, which is a real scanned dataset consisting of four point cloud sequences named Long Dress, Loot, Red and Black, and Solder respectively. Each sequence is generated by scanning a moving human body. We uniformly sample 5000 points for each point cloud. Then, we construct the input point cloud pairs in turn based on these processed point cloud frames.

However, to the best of our knowledge, the real scanned dynamic point cloud sequence dataset usually lacks ground truth information about point matches. Thus, we adopt another evaluation metric here based on the Chamfer distance. Specifically, we estimate the corresponding point for each point of the source point cloud from the target point cloud. To quantitatively compare ours with baselines, we calculate the Chamfer distance between the target point cloud and the corresponding points of the source point cloud. Formally, we notate the source and target point clouds as $\mathbf{X}$ and $\mathbf{Y}$, and the corresponding point set of $\mathbf{X}$ as $\tilde{\mathbf{X}}$. Then, the Chamfer distance $\textbf{CD}$ between $\tilde{\mathbf{X}}$ and $\mathbf{Y}$ is calculated by
\begin{equation}
\textbf{CD}(\Tilde{\mathbf{X}},\mathbf{Y}) = 
 \frac{1}{N_{\Tilde{\mathbf{X}}}} \sum \limits_{\Tilde{\mathbf{x}} \in \Tilde{\mathbf{X}}} \min \limits_{\mathbf{y}\in \mathbf{Y}} \|\Tilde{\mathbf{x}}-\mathbf{y}\|_2^2
 + \frac{1}{N_\mathbf{Y}} \sum \limits_{\mathbf{y}\in \mathbf{Y}} \min \limits_{\Tilde{\mathbf{x}}\in \Tilde{\mathbf{X}}} \|\Tilde{\mathbf{x}}-\mathbf{y}\|_2^2,
\end{equation}
where $N_{\Tilde{\mathbf{X}}}$ and $N_\mathbf{Y}$ indicate the numbers of points in the $\tilde{\mathbf{X}}$ and $\mathbf{Y}$ respectively.
Finally, we report the average Chamfer distance.

\begin{figure*}[!h]
	\centerline{\includegraphics[width=0.9\linewidth]{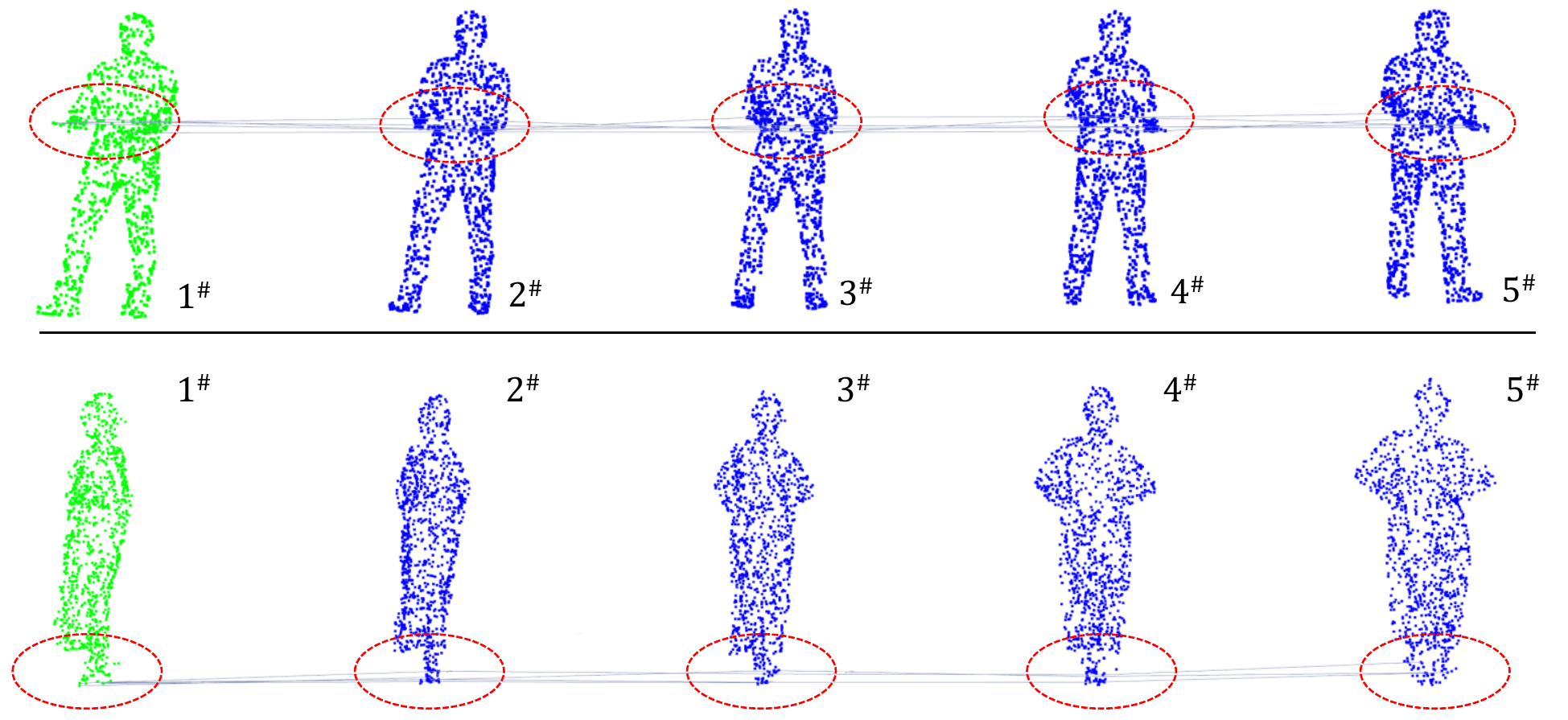}}
	\caption{Matching results for handing point cloud sequences in Loot and Long Dress from 8iVFB dataset \cite{eon_8ivfb_2017}. Here, we take the Loot as an example for a clear explanation. $1^\sharp$ indicates the frame index of the first frame. For the first frame, we select some points in the hand part, notated as $\mathcal{X}_1$. Then, we solve the corresponding points of $\mathcal{X}_1$ in the second point cloud frame, notated as $\hat{\mathcal{X}}_2$. Next, we take this second frame as the source point cloud and the third frame as the target point cloud. And according to these corresponding points $\hat{\mathcal{X}}_2$ in the second frame, we estimate the corresponding points of $\hat{\mathcal{X}}_2$ in the third frame, notated as $\hat{\mathcal{X}}_3$. This process continues in these consecutive frames. Finally, we find that all the corresponding points in consecutive frames, \ie ${\mathcal{X}}_1, \hat{\mathcal{X}}_2, \hat{\mathcal{X}}_3, ...$, keep stability in the hand part. 
	}
	\label{fig:consequence}
	\vspace{-0.3cm}
\end{figure*}

As reported in \tabref{tab:ivfb}, we can find that our EDFNet achieves the best results among all baselines, which validates the matching effectiveness of our method. To further present the temporal consistency of our method for handling the point cloud sequence, we visualize the matching results of some consecutive point cloud frames of 8iVFB dataset \cite{eon_8ivfb_2017} in \figref{fig:consequence}. To clearly observe the temporal consistency of results, we just show the matching of a local part, (\eg the hands part of the human in Loot sequence and the foot part of the human in Long Dress sequence). Here, we take the Loot as an example for a clear explanation. Specifically, for the first frame, we select some points of hand, notated as $\mathcal{X}_1$. Then, we solve the corresponding points of $\mathcal{X}_1$ in the second point cloud frame, notated as $\hat{\mathcal{X}}_2$. Next, we take this second frame as the source point cloud and the third frame as the target point cloud. And according to these corresponding points $\hat{\mathcal{X}}_2$ in the second frame, we estimate the corresponding points of $\hat{\mathcal{X}}_2$ in the third frame, notated as $\hat{\mathcal{X}}_3$. 
This process continues in these consecutive frames. Finally, we find that all the corresponding points in these consecutive frames, \ie ${\mathcal{X}}_1, \hat{\mathcal{X}}_2, \hat{\mathcal{X}}_3, ...$, keep stability in the hand part. This result validates the temporal consistency of ours for handling the point cloud sequence.

\begin{table}[t]
	\renewcommand\arraystretch{1.0}
	\caption{Average Chamfer distance comparison on the 8iVFB dataset \cite{eon_8ivfb_2017} involving four point cloud sequences.}
	\vspace{-0.2cm}
	\begin{center}
	    \resizebox{0.9\linewidth}{!}{
			\begin{tabular}{l|c|c|c|c}
				\toprule
			    \textbf{Methods}&\textbf{Long Dress} &\textbf{Loot}&\textbf{Red and Black}&\textbf{Soldier} \\
				\midrule
				D3Feat \cite{D3Feat_bai_cvpr_2020}     &0.041     &0.033  &0.037  &0.055  \\
				PointNet \cite{charles_pointnet_cvpr_2017}  &0.064     &0.057  &0.061  &0.072  \\
			   S-CorrNet3D \cite{zeng_corrnet3d_cvpr_2021}&0.014     &0.012  &0.012  &0.023\\
			   U-CorrNet3D \cite{zeng_corrnet3d_cvpr_2021}&0.057     &0.051  &0.055  &0.063\\
				S-F3D \cite{liu_flownet3d_cvpr_2019}     &0.019     &0.017  &0.018  &0.025  \\
				U-F3D \cite{liu_flownet3d_cvpr_2019}     &0.061     &0.052  &0.054  &0.066  \\
				SpinNet \cite{Ao_spinnet_cvpr_2021}   &0.022     &0.019  &0.021  &0.029  \\
				KPConv \cite{thomas_kpconv_iccv_19}     &0.034     &0.027  &0.031  &0.047  \\
				DGCNN \cite{wang_dgcnn_tog_2019}      &0.047     &0.042  &0.045  &0.059  \\
				\cdashline{1-5}[2.2pt/1.2pt]
				{EDFNet}       &\textbf{0.012}     &\textbf{0.010}  &\textbf{0.010}  &\textbf{0.019}  \\
				\bottomrule
			\end{tabular}}
	        \label{tab:ivfb}
	\end{center}
	\vspace{-0.8cm}
\end{table}	

\subsection{Ablation study}
Here, we provide several important ablation studies to analyze our proposed method. The experiments are conducted on the 3DMatch dataset unless otherwise explained.

\noindent\textbf{Performance under different numbers of selected points.} As mentioned above, the results of our evaluations are counted in some selected 3D points following D3Feat \cite{D3Feat_bai_cvpr_2020}. Here, the performance under different numbers of selected points captured by two different strategies (\ie \texttt{rand} and \texttt{pred}) is presented, where the number of selected points is reduced from 5000 to 2500, 1000, 500, and even 250. The final results are presented in \tabref{tab:3dmatch:all}. We can find that, when the points number gets smaller, the performance of PerfectMatch, FCGF, and D3Feat(rand) drops at a similar magnitude while EDFNet(rand) keeps stability and is better than others.
When the detector is equipped, EDFNet(pred) achieves the best matching quality results under all different numbers of selected points on both feature matching recall and registration recall metrics.
As for the metric of inlier ratio, our proposed EDFNet also achieves better performance in most settings compared with the \textit{SOTA} methods. These results validate that our method makes corresponding points more matchable by learning a matching task-specific feature descriptor.

\begin{table}[!t]
	\renewcommand\arraystretch{1.0}
	\caption{Performance comparison under different numbers of selected points.}
	\vspace{-0.2cm}
	\begin{center}
	    \resizebox{0.8\linewidth}{!}{
			\begin{tabular}{l|ccccc}
				\toprule
			    \textbf{Methods}&\textbf{5000}&\textbf{2500}&\textbf{1000}&\textbf{500}&\textbf{250} \\
				\midrule
				\multicolumn{6}{c}{\texttt{Feature Matching Recall} (\%)} \\
				\midrule
				{PerferctMatch \cite{perfect_zan_cvpr_2019}}  &94.7  &94.2  &92.6  &90.1  & 82.9  \\
				{FCGF \cite{choy_fcgf_iccv_2019}}             &95.2  &95.5  &94.6  &93.0  & 89.9  \\
				{D3Feat(rand) \cite{D3Feat_bai_cvpr_2020}}    &95.3  &95.1  &94.2  &93.6  & 90.8  \\
				{D3Feat(pred) \cite{D3Feat_bai_cvpr_2020}}    &95.8  &95.6  &94.6  &94.3  & 93.3  \\
				\cdashline{1-6}[2.2pt/1.2pt] 
				{EDFNet(rand) }  &95.8&95.6&{95.3}&\textbf{96.2}&93.3\\
				{EDFNet(pred) }  &\textbf{97.5}&\textbf{96.5}&\textbf{95.7}&\textbf{96.2}&\textbf{95.5}\\ %
				
				\midrule
				\multicolumn{6}{c}{\texttt{Registration Recall} (\%)} \\
				\midrule
				{PerferctMatch \cite{perfect_zan_cvpr_2019}}  &80.3  &77.5  &73.4  &64.8  & 50.9  \\
				{FCGF \cite{choy_fcgf_iccv_2019}}             &{87.3}  &85.8  &85.8  &81.0  & 73.0  \\
				{D3Feat(rand) \cite{D3Feat_bai_cvpr_2020}}    &83.5  &82.1  &81.7  &77.6  & 68.8  \\
				{D3Feat(pred) \cite{D3Feat_bai_cvpr_2020}}    &82.2  &84.4  &84.9  &82.5  & 79.3  \\
				\cdashline{1-6}[2.2pt/1.2pt] 
				{EDFNet(rand) }  &87.2&85.9&85.5&81.3&73.3\\
				{EDFNet(pred) }  &\textbf{87.5}&\textbf{86.2}&\textbf{87.9}&\textbf{87.9}&\textbf{86.3}\\
				
				\midrule
				\multicolumn{6}{c}{\texttt{Inlier Ratio} (\%)} \\
				\midrule
				{PerferctMatch \cite{perfect_zan_cvpr_2019}}  &37.7  &34.5  &28.3  &23.0  & 19.1  \\
				{FCGF \cite{choy_fcgf_iccv_2019}}             &\textbf{56.9}  &\textbf{54.5}  &49.1  &43.3  & 34.7  \\
				{D3Feat(rand) \cite{D3Feat_bai_cvpr_2020}}    &40.6  &38.3  &33.3  &28.6  & 23.5  \\
				{D3Feat(pred) \cite{D3Feat_bai_cvpr_2020}}    &40.7  &40.6  &42.7  &44.1  & 45.0  \\
				\cdashline{1-6}[2.2pt/1.2pt] 
				{EDFNet(rand) }  &52.8&51.0&45.3&43.1&39.7\\
				{EDFNet(pred) }  &55.2&\textbf{54.5}&\textbf{51.7}&\textbf{49.3}&\textbf{48.8}\\ %
				\bottomrule
			\end{tabular}}
	        \label{tab:3dmatch:all}
	\end{center}
	\vspace{-0.8cm}
\end{table}

\begin{table}[!h]
	\renewcommand\arraystretch{1.0}
	\caption{Comparison of the entire EDFNet, the EDFNet pipeline without dynamic fusion (DF) but with addition (Add) or concatenation (Cat), and the EDFNet without Transformer.}
	\vspace{-0.2cm}
	\begin{center}
	    \resizebox{0.85\linewidth}{!}{
			\begin{tabular}{l|cccccc}
				\toprule
			    \textbf{Methods}&\textbf{5000}&\textbf{2500}&\textbf{1000}&\textbf{500}&\textbf{250}&~ \\
				\midrule
				\multicolumn{7}{c}{\texttt{Feature Matching Recall} (\%)} \\
				\midrule
				{EDFNet w/o DF, w/ Add}    &93.2  &93.2  &92.6  &92.1  & 89.9 &\multirow{4}{*}{\rotatebox{90}{Rand}}  \\
				{EDFNet w/o DF, w/ Cat}    &92.9  &93.0  &92.7  &91.7  & 89.4 &~\\
				{EDFNet w/o Transformer}  &93.4  &93.3  &92.9  &93.0  & 90.3  &~ \\
				{EDFNet}                  &\textbf{95.8}  
				                         &\textbf{95.6}  
				                         &\textbf{95.3}  
				                         &\textbf{96.2}  
				                         &\textbf{93.3}  &~\\
				\cdashline{1-7}[2.2pt/1.2pt] 
				{EDFNet w/o DF, w/ Add}    &94.3  &94.1  &90.9  &92.6  & 92.8 
				&\multirow{4}{*}{\rotatebox{90}{Pred}}\\
				{EDFNet w/o DF, w/ Cat}    &94.1  &93.2  &92.2  &91.8  & 90.9 &~ \\
				{EDFNet w/o Transformer}  &94.7  &94.2  &93.0  &93.4  & 92.9  &~ \\
				{EDFNet}                  &\textbf{97.5}  
				                         &\textbf{96.5}  
				                         &\textbf{95.7}  
				                         &\textbf{96.2}  
				                         &\textbf{95.5}  &~\\
				
				\midrule
				\multicolumn{7}{c}{\texttt{Registration Recall} (\%)} \\
				\midrule
				{EDFNet w/o DF, w/ Add}    &83.7  &82.4  &81.4  &78.7  & 69.4 &\multirow{4}{*}{\rotatebox{90}{Rand}}  \\
				{EDFNet w/o DF, w/ Cat}    &83.2  &83.1  &81.6  &79.1  & 70.3 &~ \\
				{EDFNet w/o Transformer}  &83.9  &81.2  &82.6  &78.2  & 70.9  &~ \\
				{EDFNet}                  &\textbf{87.2}  
				                         &\textbf{85.9}  
				                         &\textbf{85.5}  
				                         &\textbf{81.3}  
				                         &\textbf{73.3}  &~\\
				\cdashline{1-7}[2.2pt/1.2pt] 
				{EDFNet w/o DF, w/ Add}    &83.5  &83.2  &83.3  &81.7  & 82.1  &\multirow{4}{*}{\rotatebox{90}{Pred}}\\
				{EDFNet w/o DF, w/ Cat}    &82.7  &83.1  &82.6  &81.3  & 81.2 &~ \\
				{EDFNet w/o Transformer}  &84.1  &82.9  &84.2  &82.8  & 83.2  &~ \\
				{EDFNet}                  &\textbf{87.5}  
				                         &\textbf{86.2}  
				                         &\textbf{87.9}  
				                         &\textbf{87.9}  
				                         &\textbf{86.3}  &~\\
				
				\midrule
				\multicolumn{7}{c}{\texttt{Inlier Ratio} (\%)} \\
				\midrule
				{EDFNet w/o DF, w/ Add}    &48.7  &47.8  &41.7  &39.9  & 35.7 &\multirow{4}{*}{\rotatebox{90}{Rand}}  \\
				{EDFNet w/o DF, w/ Cat}    &47.2  &46.9  &40.3  &40.1  & 36.4 &~ \\
				{EDFNet w/o Transformer}  &47.3  &47.1  &40.6  &38.4  & 36.4  &~ \\
				{EDFNet}                  &\textbf{52.8}  
				                         &\textbf{51.0}  
				                         &\textbf{45.3}  
				                         &\textbf{43.1}  
				                         &\textbf{39.7}  &~\\
				\cdashline{1-7}[2.2pt/1.2pt] 
				{EDFNet w/o DF, w/ Add}    &49.1  &47.7  &47.0  &45.6  & 44.8  &\multirow{4}{*}{\rotatebox{90}{Pred}}\\
				{EDFNet w/o DF, w/ Cat}    &48.3  &46.9  &44.3  &44.1  & 42.5 &~ \\
				{EDFNet w/o Transformer}  &48.4  &48.1  &46.6  &46.1  & 43.9  &~ \\
				{EDFNet}                  &\textbf{55.2}  
				                         &\textbf{54.5}  
				                         &\textbf{51.7}  
				                         &\textbf{49.3}  
				                         &\textbf{48.8}  &~\\
				\bottomrule
			\end{tabular}}
	        \label{tab:3dmatch:ab}
	\end{center}
	\vspace{-0.9cm}
\end{table}

\noindent\textbf{Effectiveness of each component.}
Here, we conduct several experiments on the 3DMatch dataset to analyze the effectiveness of each component of our EDFNet.
Specifically, to verify the Transformer branch in our encoder, we compare our entire framework with the results returned by learning the source and target independently without Transformer. 
Besides, to validate the importance of the proposed dynamic fusion module, we compare ours with the pipeline without dynamic fusion including two versions. 1) The decoder proposed in \cite{D3Feat_bai_cvpr_2020} is used here, which adds the different scale features together for the final feature descriptor. 2) The decoder proposed in \cite{wang_dgcnn_tog_2019} is tested, where these different scale features are concatenated for the final feature descriptor.
The results are reported in \tabref{tab:3dmatch:ab}, where other settings are totally the same for a fair comparison. Obviously, in both rand and pred settings with different numbers of selected points, the results are consistent. In particular, EDFNet with the entire pipeline outperforms the pipeline of ``EDFNet w/o dynamic fusion but with addition'', the pipeline of ``EDFNet w/o dynamic fusion but with concatenation'', and the pipeline of ``EDFNet w/o Transformer'' in all evaluation metrics. These results have validated the improvement caused by each component of our EDFNet sufficiently and clearly.

\begin{table}[!h]
	\renewcommand\arraystretch{1.0}
	\caption{Comparison between the entire EDFNet and the EDFNet pipeline without dynamic fusion but with one certain scale feature as the final feature descriptor instead of the fused one. The experiments are conducted on the 3DMatch dataset. Note that $1^\#$ indicates that the first scale feature with the minimal receptive field is used.}
	\vspace{-0.4cm}
	\begin{center}
	    \resizebox{0.9\linewidth}{!}{
			\begin{tabular}{l|cccccc}
				\toprule
			    \textbf{Methods}&\textbf{5000}&\textbf{2500}&\textbf{1000}&\textbf{500}&\textbf{250}&~ \\
				\midrule
				\multicolumn{7}{c}{\texttt{Feature Matching Recall} (\%)} \\
				\midrule
				{EDFNet w/ $1^\#$ feature}    &92.7  &92.7  &92.8  &92.9  & 90.4 &\multirow{6}{*}{\rotatebox{90}{Rand}}  \\
				{EDFNet w/ $2^\#$ feature}    &93.4  &92.9  &94.1  &93.0  & 91.7 &~\\
				{EDFNet w/ $3^\#$ feature}    &93.9  &93.7  &94.2  &93.7  & 92.2  &~ \\
				{EDFNet w/ $4^\#$ feature}    &93.2  &93.0  &93.5  &93.8  & 92.4  &~ \\
				{EDFNet w/ $5^\#$ feature}    &92.1  &92.2  &91.5  &91.2  & 90.7  &~ \\
				{EDFNet}                  &\textbf{95.8}  
				                         &\textbf{95.6}  
				                         &\textbf{95.3}  
				                         &\textbf{96.2}  
				                         &\textbf{93.3}  &~\\
				\cdashline{1-7}[2.2pt/1.2pt] 
				{EDFNet w/ $1^\#$ feature}    &93.2  &93.4  &91.5  &91.8  & 90.8 
				&\multirow{6}{*}{\rotatebox{90}{Pred}}\\
				{EDFNet w/ $2^\#$ feature}    &93.1  &93.3  &92.3  &92.1  & 91.9  &~ \\
				{EDFNet w/ $3^\#$ feature}    &94.6  &94.2  &93.2  &92.7  & 93.4  &~ \\
				{EDFNet w/ $4^\#$ feature}    &94.7  &93.8  &93.1  &92.9  & 93.8  &~ \\
				{EDFNet w/ $5^\#$ feature}    &93.1  &92.7  &91.8  &92.0  & 91.7  &~ \\
				{EDFNet}                  &\textbf{97.5}  
				                         &\textbf{96.5}  
				                         &\textbf{95.7}  
				                         &\textbf{96.1}  
				                         &\textbf{95.5}  &~\\
				
				\midrule
				\multicolumn{7}{c}{\texttt{Registration Recall} (\%)} \\
				\midrule
				{EDFNet w/ $1^\#$ feature}    &81.9  &80.8  &78.1  &74.3  & 66.2 &\multirow{6}{*}{\rotatebox{90}{Rand}}  \\
				{EDFNet w/ $2^\#$ feature}    &82.3  &81.2  &81.1  &77.3  & 70.1  &~ \\
				{EDFNet w/ $3^\#$ feature}    &82.4  &81.7  &80.1  &75.6  & 69.3  &~ \\
				{EDFNet w/ $4^\#$ feature}    &80.4  &81.1  &79.5  &71.8  & 68.4  &~ \\
				{EDFNet w/ $5^\#$ feature}    &80.2  &78.5  &72.7  &70.7  & 67.0  &~ \\
				{EDFNet}                  &\textbf{87.2}  
				                         &\textbf{85.9}  
				                         &\textbf{85.5}  
				                         &\textbf{81.3}  
				                         &\textbf{73.3}  &~\\
				\cdashline{1-7}[2.2pt/1.2pt] 
				{EDFNet w/ $1^\#$ feature}    &82.8  &80.9  &80.4  &79.3  & 77.8  &\multirow{6}{*}{\rotatebox{90}{Pred}}\\
				{EDFNet w/ $2^\#$ feature}    &82.1  &81.3  &81.1  &81.0  & 78.4  &~ \\
				{EDFNet w/ $3^\#$ feature}    &83.7  &81.2  &81.0  &81.3  & 79.9  &~ \\
				{EDFNet w/ $4^\#$ feature}    &81.8  &79.7  &80.5  &80.7  & 78.2  &~ \\
				{EDFNet w/ $5^\#$ feature}    &80.3  &78.4  &78.3  &75.2  & 75.7  &~ \\
				{EDFNet}                  &\textbf{87.5}  
				                         &\textbf{86.2}  
				                         &\textbf{87.9}  
				                         &\textbf{87.9}  
				                         &\textbf{86.3}  &~\\
				
				\midrule
				\multicolumn{7}{c}{\texttt{Inlier Ratio} (\%)} \\
				\midrule
				{EDFNet w/ $1^\#$ feature}    &44.2  &43.0  &36.3  &34.2  & 31.7 &\multirow{6}{*}{\rotatebox{90}{Rand}}  \\
				{EDFNet w/ $2^\#$ feature}    &44.9  &44.2  &38.4  &35.7  & 33.4  &~ \\
				{EDFNet w/ $3^\#$ feature}    &47.7  &46.8  &40.6  &37.8  & 35.7  &~ \\
				{EDFNet w/ $4^\#$ feature}    &45.1  &45.6  &40.0  &36.2  & 32.7  &~ \\
				{EDFNet w/ $5^\#$ feature}    &42.5  &42.0  &38.1  &35.4  & 30.9  &~ \\
				{EDFNet}                  &\textbf{52.8}  
				                         &\textbf{51.0}  
				                         &\textbf{45.3}  
				                         &\textbf{43.1}  
				                         &\textbf{39.7}  &~\\
				\cdashline{1-7}[2.2pt/1.2pt] 
				{EDFNet w/ $1^\#$ feature}    &44.7  &45.3  &43.2  &42.5  & 40.3  &\multirow{6}{*}{\rotatebox{90}{Pred}}\\
				{EDFNet w/ $2^\#$ feature}    &46.6  &47.0  &44.5  &43.8  & 41.8  &~ \\
				{EDFNet w/ $3^\#$ feature}    &49.3  &47.3  &44.1  &45.4  & 43.9  &~ \\
				{EDFNet w/ $4^\#$ feature}    &46.1  &47.4  &43.2  &44.0  & 42.4  &~ \\
				{EDFNet w/ $5^\#$ feature}    &42.7  &43.2  &43.4  &41.9  & 40.7  &~ \\
				{EDFNet}                  &\textbf{55.2}  
				                         &\textbf{54.5}  
				                         &\textbf{51.7}  
				                         &\textbf{49.3}  
				                         &\textbf{48.8}  &~\\
				\bottomrule
			\end{tabular}}
	        \label{tab:singlescale}
	\end{center}
	\vspace{-0.7cm}
\end{table}

\noindent\textbf{Single scale feature vs. multiple scale fused feature.}
As reported before, our dynamic fusion module brings better performance than handling different scale features indiscriminately and equally, \ie addition and concatenation. 
However, if we only focus on the robustness and dodging disturbers, we can use the minimal scale feature as the final descriptor rather than the fused one. 
Nevertheless, the discriminativeness is limited in this way. To further validate the effectiveness of the designed dynamic fusion module, we provide the experimental results of using each single scale feature as the final feature descriptor. The experiments are conducted on 3DMatch following the previous settings of \secref{sec::3dmatch}.
Here, according to \tabref{tab:singlescale}, we can find that the dynamic fusion of different scale features generates better performance than using any single scale feature. These results further validate our fused descriptor, which takes both robustness and discriminativeness into account to achieve better 3D point cloud matching.

\noindent\textbf{Effectiveness of the data augmentation.}
As mentioned above, during the training in our experiments, we apply data augmentation, including Gaussian jitter, random scaling, and random rotation.
To demonstrate the effectiveness of these data augmentation strategies, we provide a comparison here. Specifically, we train a model without augmentation and evaluate it on the 3DMatch dataset. Other settings are same as \secref{sec::3dmatch}. The results are presented in \tabref{supp:tab:3dmatch:aug}. We find that the results are consistent under different numbers of the selected points, and the performance of EDFNet is improved with a large margin by the augmentation operation.

\begin{table}[!h]
    \vspace{-0.2cm}
	\renewcommand\arraystretch{1.0}
	\caption{Feature match recall results on the 3DMatch dataset with and without data augmentation (aug.) under different numbers of the selected points.}
	\vspace{-0.3cm}
	\begin{center}
		\resizebox{0.75\linewidth}{!}{
			\begin{tabular}{c|ccccc}
				\toprule
				~&5000&2500&1000&500&250 \\
				\midrule				
				w/o aug. &{92.5} &{92.2} &{90.7} &{90.1} &{90.0}\\
				w/ aug.  &{\textbf{97.5}} &{\textbf{96.5}} &{\textbf{95.7}} &{\textbf{96.1}} &{\textbf{95.5}}\\
				\bottomrule
		\end{tabular}}
		\label{supp:tab:3dmatch:aug}
	\end{center}
	\vspace{-0.4cm}
\end{table}

\noindent\textbf{Contrastive loss vs. Triplet loss.}
In our method, we use metric learning with contrastive loss to train our EDFNet. Besides the contrastive loss, another typical metric learning loss function has also received widespread attention, \ie triplet loss function \cite{choy_fcgf_iccv_2019}. Here, we provide a comparison between these two loss functions. Other settings are same as \secref{sec::3dmatch}. The results are given in \tabref{supp:tab:3dmatch:loss}, where we provide the approximate epoch when the network converges besides the feature match recall results. We can find that the two metric learning loss functions lead to similar feature match recall results. Nevertheless, the contrastive loss helps EDFNet to converge much faster.

\begin{table}[!h]
	\renewcommand\arraystretch{1.0}
 	\vspace{-0.2cm}
	\caption{Comparison between the contrastive loss and triplet loss to train our EDFNet on the 3DMatch dataset. \textbf{AVG.} indicates the average feature match recall results. Epoch$^\flat$ indicates the approximate epoch when the network converges.}
	\vspace{-0.3cm}
	\begin{center}
		\resizebox{0.5\linewidth}{!}{
			\begin{tabular}{c|cc}
				\toprule
				~&\textbf{AVG.}&Epoch$^\flat$ \\
				\midrule				
				Contrastive &97.543 &100\\
				Triplet  &97.596 &170\\
				\bottomrule
		\end{tabular}}
		\label{supp:tab:3dmatch:loss}
	\end{center}
	\vspace{-0.4cm}
\end{table}

\section{Discussion and conclusion}\label{conclusion}

In this paper, we have proposed a novel \emph{task-specific} point cloud feature descriptor learning method to tackle the 3D point cloud matching problem. We has presented EDFNet to learn point cloud descriptor, \ie pruning the descriptor according to the specific task. 
Specifically, considering that the matching task aims at discovering point correspondences, we proposed to make the correspondence more matchable by exploiting the repetitive structure in two input point clouds. A CT-encoder is applied to extract different scale features. It captures the local geometry from the current point cloud by convolution and parallel searches the co-contextual information from paired point cloud by Transformer. Besides, a dynamic fusion module is designed to fuse these different scale features adaptively. It guides the final descriptor approach to the consistent and clean ones to dodge the disturbers to achieve a robust and discriminative feature descriptor. Extensive experiments on different datasets have validated the superiority of our proposed EDFNet. In the future, this task-specific learning method can be extended to other matching tasks, such as 2D-2D matching and 2D-3D matching.

However, the limitations of our method are also obvious. Since EDFNet is customized for the matching task, it is limited in other applications, such as classification, segmentation, \etc. Besides, although we combine the descriptor learning with the matching task, our method still cannot handle the similar or symmetrical structures well. This is a stubborn problem in the point cloud matching community without effective solutions \cite{D3Feat_bai_cvpr_2020,Ao_spinnet_cvpr_2021,zhang_self_pr_2022}.

\section*{Acknowledgement}
\thanks{This work was supported in part by the National Key Research and Development Program of China under Grant 2018AAA0102803 and National Natural Science Foundation of China (61871325, 61901387). This work was also sponsored by Innovation Foundation for Doctor Dissertation of Northwestern Polytechnical University (CX2022046).}

\end{document}